\PassOptionsToPackage{table}{xcolor}
\documentclass[10pt,twocolumn,letterpaper]{article}

\usepackage[pagenumbers]{iccv} 

\usepackage[dvipsnames]{xcolor}

\usepackage{lipsum}
\usepackage{graphicx}
\usepackage{float}
\usepackage{dblfloatfix}
\usepackage{forest}
\usepackage{tikz}
\usepackage{amsmath}        
\usepackage{esint}
\usepackage{array}
\usepackage{threeparttable}
\usepackage{multirow}
\usepackage{caption}
\usepackage{subcaption}
\usepackage{mathabx}
\usepackage{arydshln}
\usepackage{enumitem}
\usepackage{bigdelim}
\usepackage{makecell} 
\usepackage[basic,italic,defaultimath,nohbar,defaultmathsizes]{mathastext}

\usepackage[resetlabels]{multibib}
\newcites{supp}{References}

\usepackage{siunitx}
\definecolor{plotblue}{RGB}{0, 68, 136}
\definecolor{plotred}{RGB}{192, 0, 0}

\usetikzlibrary{shadows, shapes, positioning, arrows.meta, ext.paths.ortho, matrix, decorations.pathreplacing}

\DeclareRobustCommand{\svdots}{
  \vbox{%
    \baselineskip=0.33333\normalbaselineskip%
    \lineskiplimit=0pt
    \hbox{.}\hbox{.}\hbox{.}%
    \kern-0.2\baselineskip%
  }%
}

\definecolor{iccvblue}{rgb}{0.21,0.49,0.74}
\usepackage[pagebackref,breaklinks,colorlinks,allcolors=iccvblue,hypertexnames=false]{hyperref}

\def\paperID{12716} 
\def\confName{ICCV}
\def\confYear{2025}

\title{\vspace{-0.2in}Hierarchical Material Recognition from Local Appearance\vspace{-0.05in}}

\author{Matthew Beveridge and Shree K.\ Nayar\\
Columbia University\\
{\tt\small \{beveridge,nayar\}@cs.columbia.edu}
}

\begin{document}

\maketitle

\begin{abstract}
    We introduce a taxonomy of materials for hierarchical recognition from local appearance. Our taxonomy is motivated by vision applications and is arranged according to the physical traits of materials. We contribute a diverse, in-the-wild dataset with images and depth maps of the taxonomy classes. Utilizing the taxonomy and dataset, we present a method for hierarchical material recognition based on graph attention networks. Our model leverages the taxonomic proximity between classes and achieves state-of-the-art performance. We demonstrate the model's potential to generalize to adverse, real-world imaging conditions, and that novel views rendered using the depth maps can enhance this capability. Finally, we show the model's capacity to rapidly learn new materials in a few-shot learning setting.
\end{abstract}

\vspace{-0.05in}
\section{Importance of Material Recognition}\label{sec:introduction}

Our ability as humans to recognize materials is critical to every action we take. Using vision alone, we can infer that a coffee cup will be hot to the touch if we see it is made of paper. If it is instead perceived as a ceramic cup, we can deduce it may be cooler and more rigid, but heavier to lift. The ability to perceive materials from a distance thus enables us to infer the consequences of an action before taking it. While this is second nature to us, humans, material perception by machines remains an active field of research.

Material is a fundamental visual unit of a scene. Objects, defined by their form and function, convey \emph{what} is in a scene. In concert, materials inform \emph{how} to interact with the scene. Consider the case where our coffee cup is toppled and spills its contents. An autonomous agent must first identify the spilled material as a liquid to then reason that a towel is the proper cleanup tool, and not a broom. Furthermore, knowing a material enables estimation of mechanical properties such as weight and elasticity. In the case of our example, the agent must know that the towel is deformable in order to handle it. In short, the ability to visually identify materials is key to the development of autonomous systems that can interact with the environment in more nuanced and intelligent ways than possible today.

The term ``material'' can carry different meanings depending on the context. In our example, if the coffee is spilled on a wooden table, it may suffice to know that coffee is a liquid. However, if it is spilled on a fabric, the cleaning method used may be based on the liquid type (\eg, coffee v.\ water). In other words, material recognition may need to be done at different levels of detail based on the application~\cite{adelsonSeeingStuffPerception2001}.

Just as naturalists have taxonomized living organisms into the tree of life, materials too conform to a hierarchy (\eg, rubber \(\subset\) plastic \(\subset\) polymer). With this in mind, we constructed a hierarchical taxonomy for materials (\cref{sec:taxonomy}). Our taxonomy includes materials that are commonly encountered in a variety of science and engineering disciplines, and its hierarchical structure is based on the physical properties of the materials. We have collected an in-the-wild image dataset, \emph{Matador}, to populate the taxonomy with images of materials taken at different scales (magnifications) and under different natural lighting conditions (\cref{sec:dataset}).\footnote{\emph{Matador} webpage: \href{https://cave.cs.columbia.edu/repository/Matador}{https://cave.cs.columbia.edu/repository/Matador}.} 
\emph{Matador} has \(\sim\)7,200 samples across 57 material classes, where each sample includes the material's local appearance (close-up texture), 3D structure (from lidar), and the surrounding context (object-level information). Having 3D structure allows us to render arbitrarily many additional novel views for each sample, corresponding to different magnifications, orientations, and camera settings.

We use the taxonomy to develop a model for material recognition by framing recognition as a hierarchical image classification task. This is done by utilizing graph representation learning to predict the full taxonomic classification of a material from its local appearance (\cref{sec:method}). We show that structuring a graph neural network according to our taxonomy improves classification accuracy on existing benchmarks as well as our own \textit{Matador} dataset, achieving state-of-the-art performance.
The performance remains high even when the taxonomy is sparsely populated with images and when color information is excluded. A key advantage of our hierarchical approach is that, even when a material is misclassified at the finest level, it can be correctly recognized at a higher level, still enabling useful inferences regarding its mechanical properties. Finally, we demonstrate that our model generalizes to adverse, real-world imaging conditions, that rendering novel views for training data can enhance this capability, and that the model is effective at few-shot classification (\cref{sec:discussion}).

\section{Related Work}\label{sec:relatedwork}

Images can be interpreted at multiple levels of granularity. At the coarsest level, an image represents a scene. Within a scene, there are objects, and each object is composed of materials. Each material manifests as a texture in the image.  Although texture and material are distinct concepts, they are closely related, and much of the prior work has utilized textures to recognize materials. We refer the interested reader to~\citet{liuBoWCNNTwo2019} and~\citet{danaComputationalTexturePatterns2018} for comprehensive historical perspectives on this line of work.

\textbf{Material Recognition from Texture.} The earliest approaches to texture recognition are based on the use of filter banks~\cite{lawsRapidTextureIdentification1980,bovikMultichannelTextureAnalysis1990,jainUnsupervisedTextureSegmentation1991,turnerTextureDiscriminationGabor1986,manjunathTextureFeaturesBrowsing1996} or statistical analysis~\cite{crossMarkovRandomField1983,maoTextureClassificationSegmentation1992}. Significant progress was later made by representing a texture using textons~\cite{juleszExperimentsVisualPerception1975} that are constructed from the frequency responses to a set of hand-crafted filters~\cite{kaplanExtendedFractalAnalysis1999,malikContourTextureAnalysis2001,culaCompactRepresentationBidirectional2001,leungRepresentingRecognizingVisual2001,varmaStatisticalApproachTexture2005,varmaLocallyInvariantFractal2007,shottonSemanticTextonForests2008,cula3DTextureRecognition2004,varmaStatisticalApproachMaterial2009}, including Gabor filters~\cite{fogelGaborFiltersTexture1989,turnerTextureDiscriminationGabor1986} and Gaussian kernels~\cite{koenderinkStructureImages1984,malikContourTextureAnalysis2001}. Textons are aggregated into a histogram and used for recognition, similar to bags of features for object recognition. Subsequent methods enhanced the filter bank to handle multi-scale image features~\cite{hadjidemetriouMultiresolutionHistogramsTheir2004,debonetTextureRecognitionUsing1998,lazebnikBagsFeaturesSpatial2006}, rotation invariance~\cite{schmidConstructingModelsContentbased2001} and affine invariance~\cite{lazebnikSparseTextureRepresentation2005}, and the downstream recognition routines were improved with Fisher vectors~\cite{perronninFisherKernelsVisual2007} and contextual priors~\cite{liuExploringFeaturesBayesian2010}.

The field has since shifted from hand-crafted filter banks to learned models. Most works in this vein utilize convolutional architectures (CNNs) to either learn a filter bank~\cite{cimpoiDeepFilterBanks2015,cimpoiDescribingTexturesWild2014} or a dictionary~\cite{songLocallyTransferredFisherVectors2017,xueDeepTextureManifold2018,xueDifferentialViewpointsGround2022,zhangDeepTENTexture2017,scabiniRADAMTextureRecognition2023}. Since forming histograms from textons is effectively spatial pooling, much attention has been given to feature aggregation in CNNs~\cite{zhaiDeepMultipleAttributePerceivedNetwork2019,zhaiDeepStructureRevealedNetwork2020,yangDFAENDoubleorderKnowledge2022}. Recently, incorporating multi-scale geometry in feature pooling has shown promising results~\cite{chenDeepTextureRecognition2021,xuEncodingSpatialDistribution2021,mohanLacunarityPoolingLayers2024,florindoFractalPoolingNew2024}. Such methods aim to replace average or max pooling with an operator that captures the orderless, fine-grained structure of textures without losing the higher-level spatial order. In contrast to prior work, we use a hierarchical taxonomy of materials, which allows us to leverage the taxonomic proximity between materials for recognition.

\textbf{Material Image Datasets.} Texture-based recognition aims to classify materials based on the intensity fluctuations they produce in images. Such fluctuations can be described as having characteristic patterns such as ``striped'' or ``checkered''~\cite{cimpoiDescribingTexturesWild2014,schwartzVisualMaterialTraits2013,schwartzAutomaticallyDiscoveringLocal2015}. In this approach, there is no direct measurement of the physical properties of a material or the way it interacts with incident light. Ideally, material recognition would be explicitly based on the optical properties of a material~\cite{danaComputationalTexturePatterns2018}. However, this would require at least a partial measurement of the bidirectional reflectance distribution function (BRDF)~\cite{zhangReflectanceHashingMaterial2015} or the bidirectional texture function (BTF)~\cite{danaReflectanceTextureRealworld1999,sattlerEfficientRealisticVisualization,weinmannMaterialClassificationBased2014,caputoClassspecificMaterialCategorisation2005}, which can only be done in controlled settings using specialized setups.

Learning-based approaches to material recognition instead lean on large image datasets to capture variability in camera pose and real-world lighting. Several datasets of this type exist and are either aggregated from online sources~\cite{sharanRecognizingMaterialsUsing2013,bellMaterialRecognitionWild2015,bellOpenSurfacesRichlyAnnotated2013} or captured using a custom hardware platform~\cite{xueDifferentialAngularImaging2017,xueDeepTextureManifold2018}. The former tend to also include some object-level information which can aid local material recognition~\cite{schwartzMaterialRecognitionLocal2017}. In our work, we are chiefly interested in recognition based on local appearance. We present a novel dataset, \emph{Matador}, of \(\sim\)7,200 in-the-wild samples spanning 57 material types. Using this dataset, we also render numerous novel views of each sample to augment model training by varying magnification, orientation, and camera settings~\cite{cheesemanScaleAmbiguitiesMaterial2022}.

\textbf{Graph Representation Learning.} The definition of a material is relative: while ``brick'' and ``concrete'' are distinct classes of materials, they are both instances of a ``ceramic''. This observation inspired our use of hierarchical learning to recognize materials. Hierarchical image classification~\cite{guoCNNRNNLargescaleHierarchical2018,salakhutdinovLearningShareVisual2011,changYourFlamingoMy2021,chenFineGrainedRepresentationLearning2018,liuWhereFocusInvestigating2022,yiExploringHierarchicalGraph2022,zhaoGraphbasedHighOrderRelation2021,yamazakiHierarchicalImageClassification} has recently been used to take advantage of semantic relationships between classes, \eg, in object recognition, where the hierarchy is defined by WordNet~\cite{millerWordNetLexicalDatabase1995}. Several such methods utilize graph neural networks~\cite{xingLearningHierarchicalGraph2021,xiaHGCLIPExploringVisionLanguage2024} to explicitly represent hierarchical relationships, which have demonstrated impressive generalization capabilities~\cite{yiExploringHierarchicalGraph2022,kampffmeyerRethinkingKnowledgeGraph2019,wangZeroShotRecognitionSemantic2018,nayakZeroShotLearningCommon2022,wangZeroShotLearningContrastive2021}. Based on our taxonomy, we develop a graph neural network~\cite{kipfSemiSupervisedClassificationGraph2017} to learn visual relationships between material classes. In doing so, should related taxonomy classes share visual traits (which is not necessarily guaranteed), we can exploit this for recognition.


\section{A Visual Taxonomy of Materials}\label{sec:taxonomy}

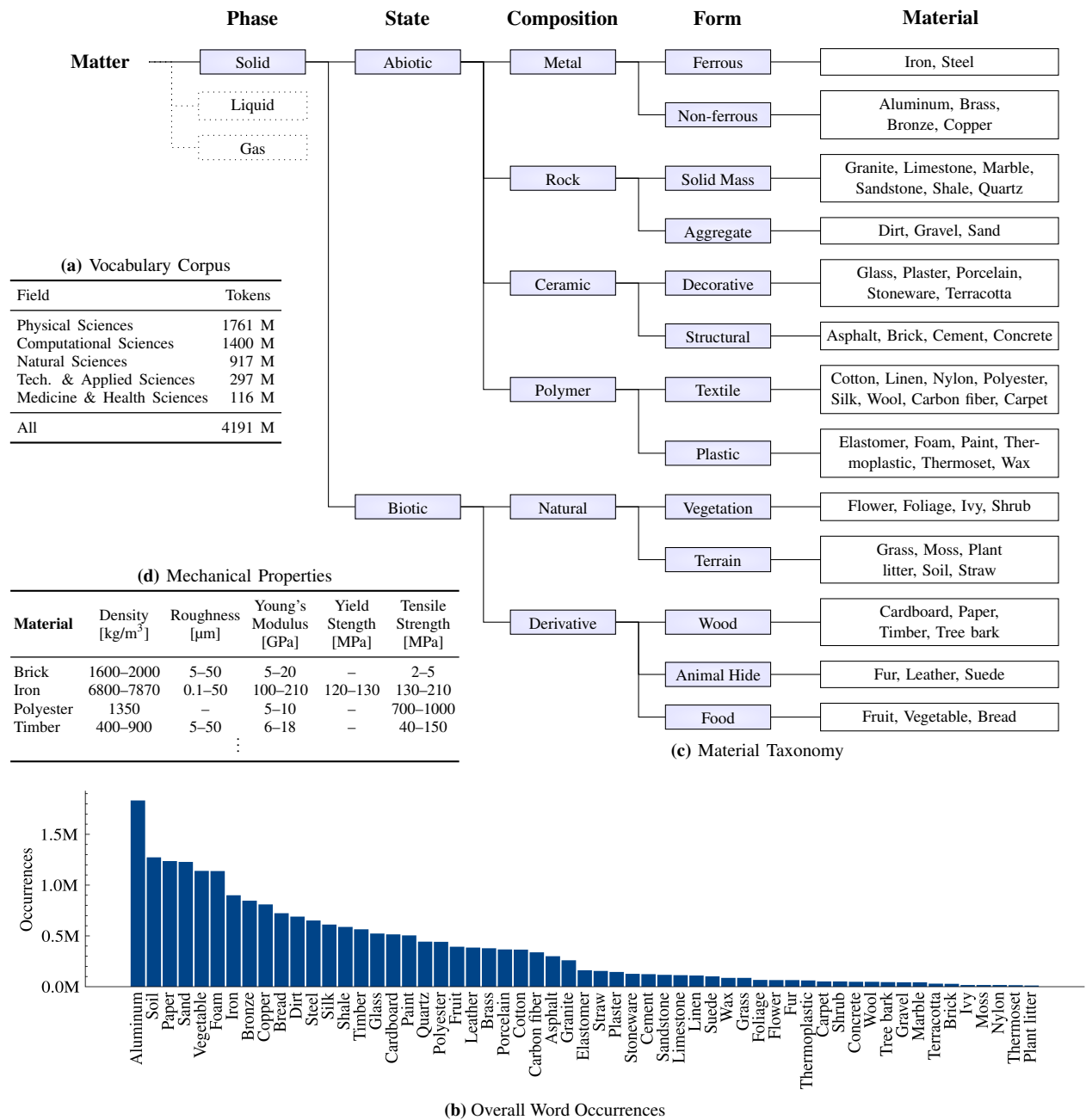
\begin{figure*}
    \vspace{1cm}
    \begin{center}
        \tikzset{
            my node/.style={
                    draw=black,
                    inner color=blue!5,
                    outer color=blue!10,
                    minimum width=3.25cm,
                    text height=2.5ex,
                    font=\Large,
                    solid
                },
            leaf node/.style={
                    text width=7cm,
                    align=center,
                    inner color=white,
                    outer color=white,
                }
        }
        \begin{forest}
            for tree={%
            my node,
            scale=0.5,
            l sep+=10pt,
            inner sep=5pt,
            edge={black},
            grow'=east,
            parent anchor=east,
            child anchor=west,
            calign=child edge,
            edge path={
                    \noexpand\path[draw, \forestoption{edge}] (!u.parent anchor) -- + (10pt,0) |- (.child anchor)\forestoption{edge label};
                },
            if n children=0{
                    leaf node
                }{
                    text depth=0ex
                },
            }
            [\textbf{Matter}, name=col0, inner color=white, outer color=white, draw=white, minimum width=3cm, font=\LARGE
            [Solid, name=col1
            [Abiotic, name=col2
            [Metal, name=col3
            [Ferrous, name=col4
            [{Iron, Steel}, name=col5]
            ]
            [Non-ferrous
            [{Aluminum, Brass, Bronze, Copper}]
            ]
            ]
            [Rock
                [Solid Mass
                        [{Granite, Limestone, Marble, Sandstone, Shale, Quartz}]
                ]
                [Aggregate
                        [{Dirt, Gravel, Sand}]
                ]
            ]
            [Ceramic
                [Decorative
                        [{Glass, Plaster, Porcelain, Stoneware, Terracotta}]
                ]
                [Structural
                        [{Asphalt, Brick, Cement, Concrete}]
                ]
            ]
            [Polymer
                [Textile
                        [{Cotton, Linen, Nylon, Polyester, Silk, Wool, Carbon fiber, Carpet}]
                ]
                [Plastic
                        [{Elastomer, Foam, Paint, Thermoplastic, Thermoset, Wax}]
                ]
            ]
            ]
            [Biotic
                [Natural
                        [Vegetation
                                [{Flower, Foliage, Ivy, Shrub}]
                        ]
                        [Terrain
                                [{Grass, Moss, Plant litter, Soil, Straw}]
                        ]
                ]
                [Derivative
                        [Wood
                                [{Cardboard, Paper, Timber, Tree bark}]
                        ]
                        [Animal Hide
                                [{Fur, Leather, Suede}]
                        ]
                        [Food
                                [{Fruit, Vegetable, Bread}]
                        ]
                ]
            ]
            ]
            [Liquid, dotted, edge={dotted}, text width=3cm]
            [Gas, dotted, edge={dotted}, text width=3cm]
            ]
        \end{forest}

        \vspace{-6mm}
        \begin{tikzpicture}[overlay]

            \node[font=\small\bfseries, above right=108mm and -78.5mm of col1, anchor=north] {Phase};
            \node[font=\small\bfseries, above right=108mm and -78.5mm of col2, anchor=north] {State};
            \node[font=\small\bfseries, above right=108mm and -78.5mm of col3, anchor=north] {Composition};
            \node[font=\small\bfseries, above right=108mm and -78.5mm of col4, anchor=north] {Form};
            \node[font=\small\bfseries, above right=108mm and -88.75mm of col5, anchor=north] {Material};

            \node[font=\small, below right=-75mm and -97mm of col0, anchor=north west, text width=5cm, align=center] (sources) {
                \subcaption{Vocabulary Corpus}\label{fig:taxonomy-corpus}
                {\scriptsize
                    \setlength{\tabcolsep}{3pt}
                    \begin{tabular}{lr}
                        \toprule
                        Field                       & \multicolumn{1}{c}{Tokens} \\
                        \midrule
                        Physical Sciences           & 1761~M                     \\
                        Computational Sciences      & 1400~M                     \\
                        Natural Sciences            & 917~M                      \\
                        Tech. \& Applied Sciences   & 297~M                      \\
                        Medicine \& Health Sciences & 116~M                      \\

                        \midrule
                        All                         & 4191~M                     \\
                        \bottomrule
                    \end{tabular}
                }
            };
            \node[font=\small, below right=-27mm and -93mm of col0, anchor=north west, text width=7cm, align=center] (mech-properties) {
                \addtocounter{subfigure}{2}
                \subcaption{Mechanical Properties}\label{fig:taxonomy-properties}
                \addtocounter{subfigure}{-2}
                {\scriptsize
                    \setlength{\tabcolsep}{1.925pt}
                    \begin{tabular}{m{9mm}
                        >{\centering}m{12mm}
                        >{\centering}m{13mm}
                        >{\centering}m{9mm}
                        >{\centering}m{9mm}
                        >{\centering\arraybackslash}m{10mm}}
                        \toprule
                        \textbf{Material} & Density [\unit{\kilogram\per\meter\cubed}] & Surface Roughness [\unit{\um}] & Young's Modulus [\unit{\GPa}] & Yield Strength [\unit{\MPa}] & Tensile Strength [\unit{\MPa}] \\ 
                        \midrule
                        Brick             & 1600--2000                                 & 5--50                          & 5--20                         & --                           & 2--5                           \\ 
                        Iron              & 7150--7870                                 & 0.1--50                        & 100--210                      & 120--200                     & 130--210                       \\ 
                        Polyester         & 1330--1380                                 & 0.004--0.006                   & 3--7                          & --                           & 200--500                       \\ 
                        Timber            & 400--900                                   & 5--50                          & 6--18                         & 25--140                      & 40--150                        \\ 
                        \multicolumn{6}{c}{\svdots}                                                                                                                                                                     \\ 
                        \bottomrule
                    \end{tabular}
                }
            };

            \node[font=\small, below right=7.75mm and 44mm of col0, anchor=south east, text width=5cm, align=center] {
                \subcaption{Material Taxonomy}
                \label{fig:taxonomy-tree}
            };

        \end{tikzpicture}

        \vspace{10mm}

        \includegraphics[width=0.95\textwidth, height=4.75cm]{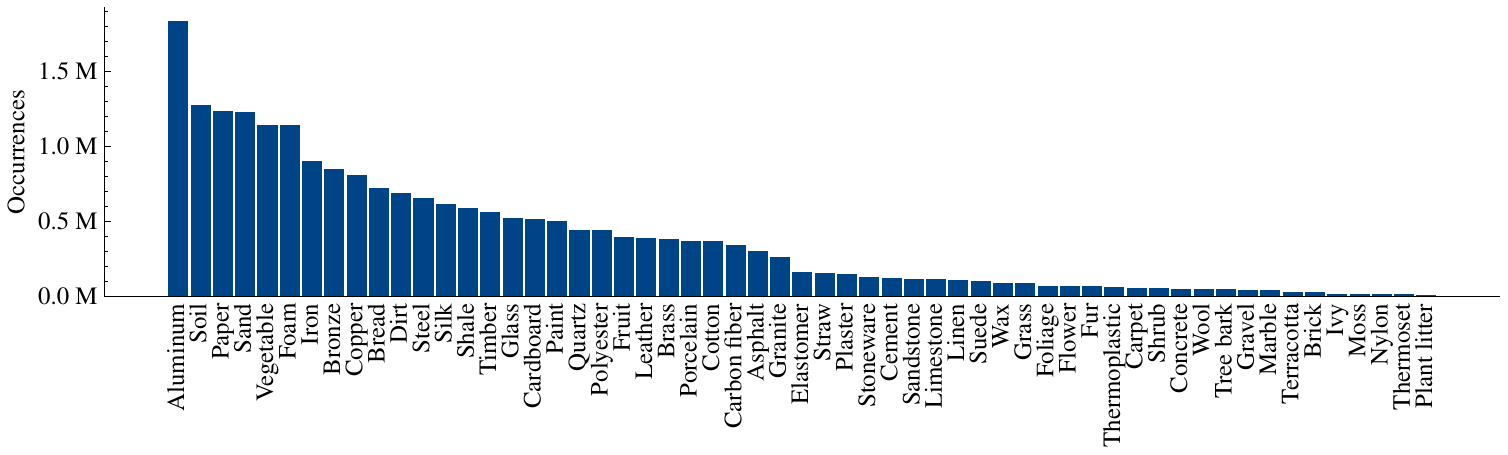}
        {\small
            \addtocounter{subfigure}{-2}
            \subcaption{Overall Word Occurrences}\label{fig:taxonomy-occurrences}
            \addtocounter{subfigure}{2}
        }

    \end{center}

    \caption{
        \textbf{Framework for the Taxonomic Classification of Materials.} (a)~We search through archives of text in the scientific literature to build a vocabulary for materials based on the frequency of word occurrences. (b)~Shown here is the word occurrence plot for the 57 materials that are most popular in the fields listed in (a). Such a vocabulary can be generated for subfields as well (\eg, robotics). (c)~Using the vocabulary in (b), we create a taxonomy for the classification of materials. The structure of the taxonomy is organized based on the physical traits shared by the materials. The advantage of this taxonomy in the context of recognition is that, even when a material cannot be uniquely identified from its appearance, it still may be identified at a higher level in the taxonomy (form, composition, state, or phase). Note that the taxonomy focuses on solids, but could be extended to liquids and gases as well in the future. (d)~In some applications, it may be useful to know how a material would behave when physically interacted with. To this end, we have listed the mechanical properties for the materials in our taxonomy (the full list is in \cref{sec:sup-material-properties} of the supplemental material).
    }\label{fig:taxonomy}
\end{figure*}

The names given to material categories are highly dependent on the context in which they are studied. While a roboticist may broadly characterize a cup as being made of plastic, a material scientist is likely to define it more finely, perhaps as polystyrene~\cite{jrCallistersMaterialsScience2020}. This is in part because different fields study materials at different scales. At one end, a material could be coarsely defined as a solid, and at the other end as a group of atoms. The granularity at which a material is perceived, therefore, depends on the level at which we wish to interact with it. We take inspiration from the tree of life in biology and define a taxonomy of materials, where the vocabulary is chosen to be pertinent to tasks requiring visual perception, and the structure is designed based on the physical traits of the materials.\footnote{Modern phylogenetic trees of life are rooted in genome sequencing. Before this, most trees of life were based on external morphology (\ie, visual appearance of organisms). In some ways, our approach represents a blend of both methods: our material taxonomy is organized by physical properties, but our recognition method utilizes passive visual imaging.} The purpose of this taxonomy is to enable the creation of a framework that can (a)~exploit visual similarities among related categories to improve classification, (b)~relate unknown materials to known ones, and (c)~make qualitative inferences about the mechanical properties of a material from its image.

\textbf{A Vocabulary of Materials.}\label{sec:vocabulary}
Inspired by \citet{bhushanTextureLexiconUnderstanding1997}, we first set out to create a vocabulary that includes all types of materials an intelligent system might encounter. We began with all the names of materials in WordNet~\cite{millerWordNetLexicalDatabase1995}. We then scanned for occurrences of these names in historical corpora of text, ranging in focus from popular science to niche conference and journal publications. The fields we focused on are listed in \cref{fig:taxonomy-corpus}. The complete corpus has billions of words and phrases sourced from M2D2~\cite{reidM2D2MassivelyMultidomain2022}. We then aggregated word occurrences of synonyms, and each synonym group is given a material name. We end up with a distilled set of 57 categories, the word occurrence distribution for which is shown in \cref{fig:taxonomy-occurrences}. One could use all of these categories, or, for a given application domain, one could regenerate the distribution of word occurrences to find the categories that are most relevant to that domain.

\textbf{Hierarchical Representation.}
Next, we arranged the above material categories into a taxonomy. Our taxonomy, inspired by \citet{schwartzRecognizingMaterialProperties2020}, is shown in \cref{fig:taxonomy-tree}. The higher-level structure (phase, state, composition, and form) was derived from existing material classifications used in the fields of material science, mechanical engineering, and chemical engineering. Note that we focus on solids in our work as they are most relevant to vision applications such as robotics and autonomous driving. Furthermore, the definitions of some of the materials have been stretched to accommodate corner cases. This is because most engineered materials are made of multiple constituent materials. In such cases, we deemed it more useful to categorize by the dominant component. This taxonomy is intended to serve as a starting point, and it can be expanded (to include liquids, for instance) with time.

\textbf{Mechanical Properties.}
For an intelligent system to fully benefit from material recognition, it must know more than just the names of the materials in its field of view. It must also have an understanding of how each material will behave upon interaction. To this end, we have compiled a table of mechanical properties for each leaf (material) in our taxonomy. The relevant parameters include density, surface roughness, elasticity, and strength. The ranges for these parameters were obtained from existing literature in material science and mechanical engineering. An abbreviated version of this table is shown in \cref{fig:taxonomy-properties}; the full table with all the materials is in the supplemental material (\cref{sec:sup-material-properties}).

\section{\emph{Matador}: A Material Image Dataset}\label{sec:dataset}

\begin{figure}
    \centering
    \footnotesize
    \setlength{\tabcolsep}{1pt}
    \begin{tabular}{m{2.5mm}p{\dimexpr0.98\columnwidth-2.5mm\relax}}
                                                      &
        \begin{minipage}{\linewidth}
            \begin{subfigure}[b]{0.24\linewidth}
                \caption{\centering Local Appearance}\label{fig:dataset-mosaic-color}
            \end{subfigure}
            \hfill
            \begin{subfigure}[b]{0.24\linewidth}
                \caption{\centering B\&W Local Appearance}\label{fig:dataset-mosaic-grayscale}
            \end{subfigure}
            \hfill
            \begin{subfigure}[b]{0.24\linewidth}
                \caption{\centering Lidar Local 3D Structure}\label{fig:dataset-mosaic-depth}
            \end{subfigure}
            \hfill
            \begin{subfigure}[b]{0.24\linewidth}
                \caption{\centering Surrounding Global Context}\label{fig:dataset-mosaic-context}
            \end{subfigure}
        \end{minipage}                     \\

        \centering \rotatebox{90}{\textbf{Brick}}     &
        \begin{minipage}{\linewidth}
            \centering
            \begin{subfigure}[b]{0.24\linewidth}
                \includegraphics[width=\linewidth, height=\linewidth]{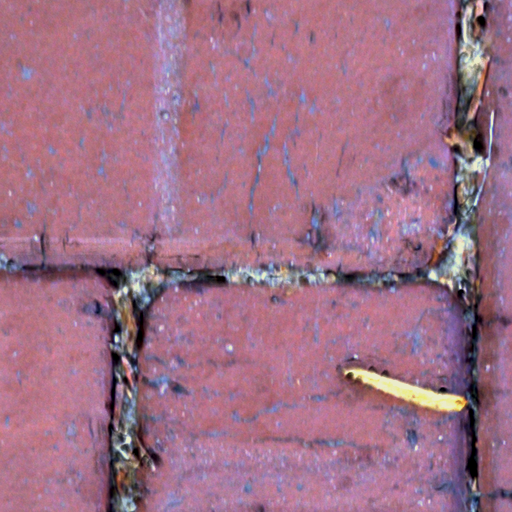}
            \end{subfigure}
            \hfill
            \begin{subfigure}[b]{0.24\linewidth}
                \includegraphics[width=\linewidth, height=\linewidth]{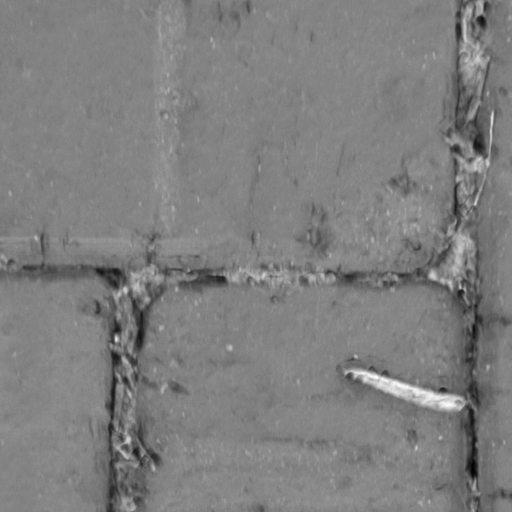}
            \end{subfigure}
            \hfill
            \begin{subfigure}[b]{0.24\linewidth}
                \includegraphics[width=\linewidth, height=\linewidth]{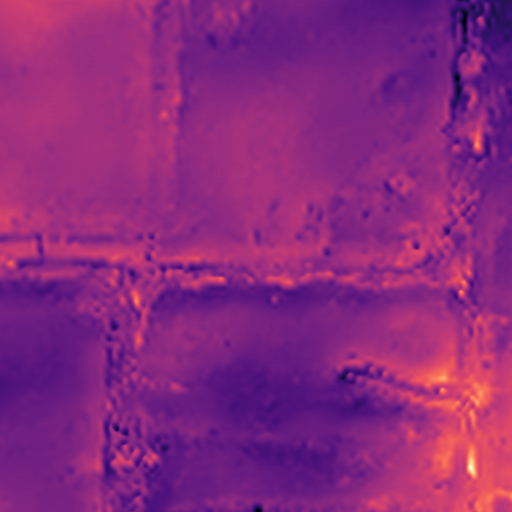}
            \end{subfigure}
            \hfill
            \begin{subfigure}[b]{0.24\linewidth}
                \includegraphics[width=\linewidth, height=\linewidth]{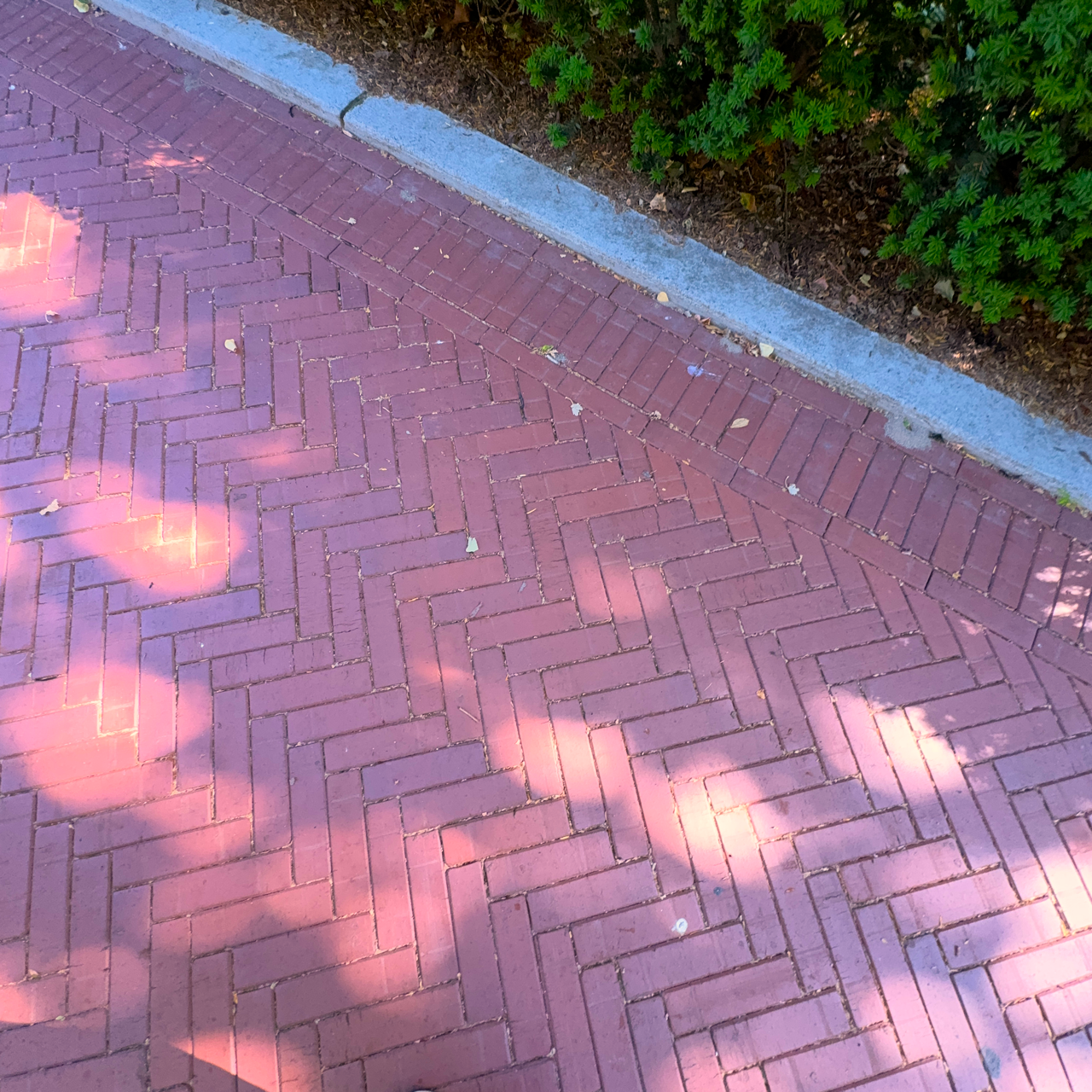}
            \end{subfigure}
            \vspace{2pt}
        \end{minipage}                     \\

        \centering \rotatebox{90}{\textbf{Foliage}}   &
        \begin{minipage}{\linewidth}
            \centering
            \begin{subfigure}[b]{0.24\linewidth}
                \includegraphics[width=\linewidth, height=\linewidth]{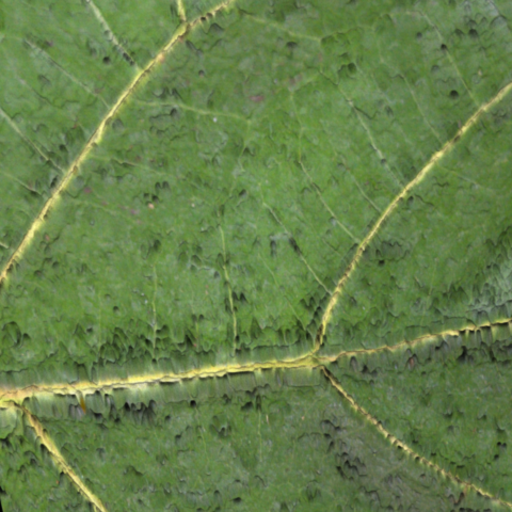}
            \end{subfigure}
            \hfill
            \begin{subfigure}[b]{0.24\linewidth}
                \includegraphics[width=\linewidth, height=\linewidth]{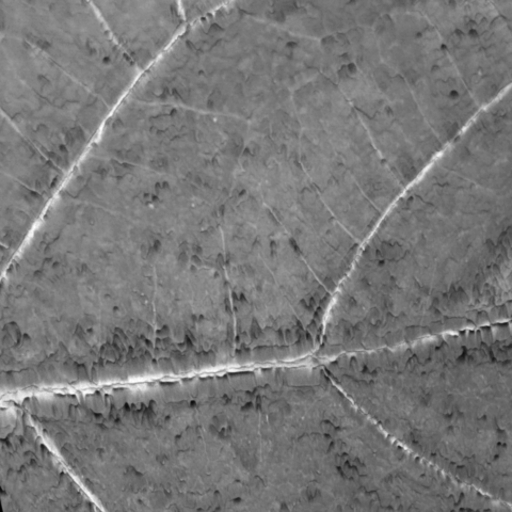}
            \end{subfigure}
            \hfill
            \begin{subfigure}[b]{0.24\linewidth}
                \includegraphics[width=\linewidth, height=\linewidth]{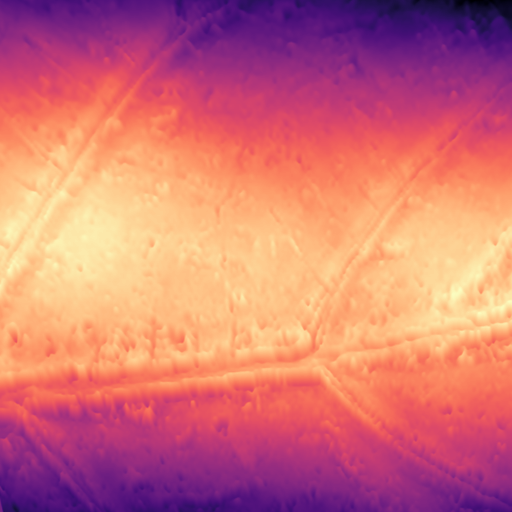}
            \end{subfigure}
            \hfill
            \begin{subfigure}[b]{0.24\linewidth}
                \includegraphics[width=\linewidth, height=\linewidth]{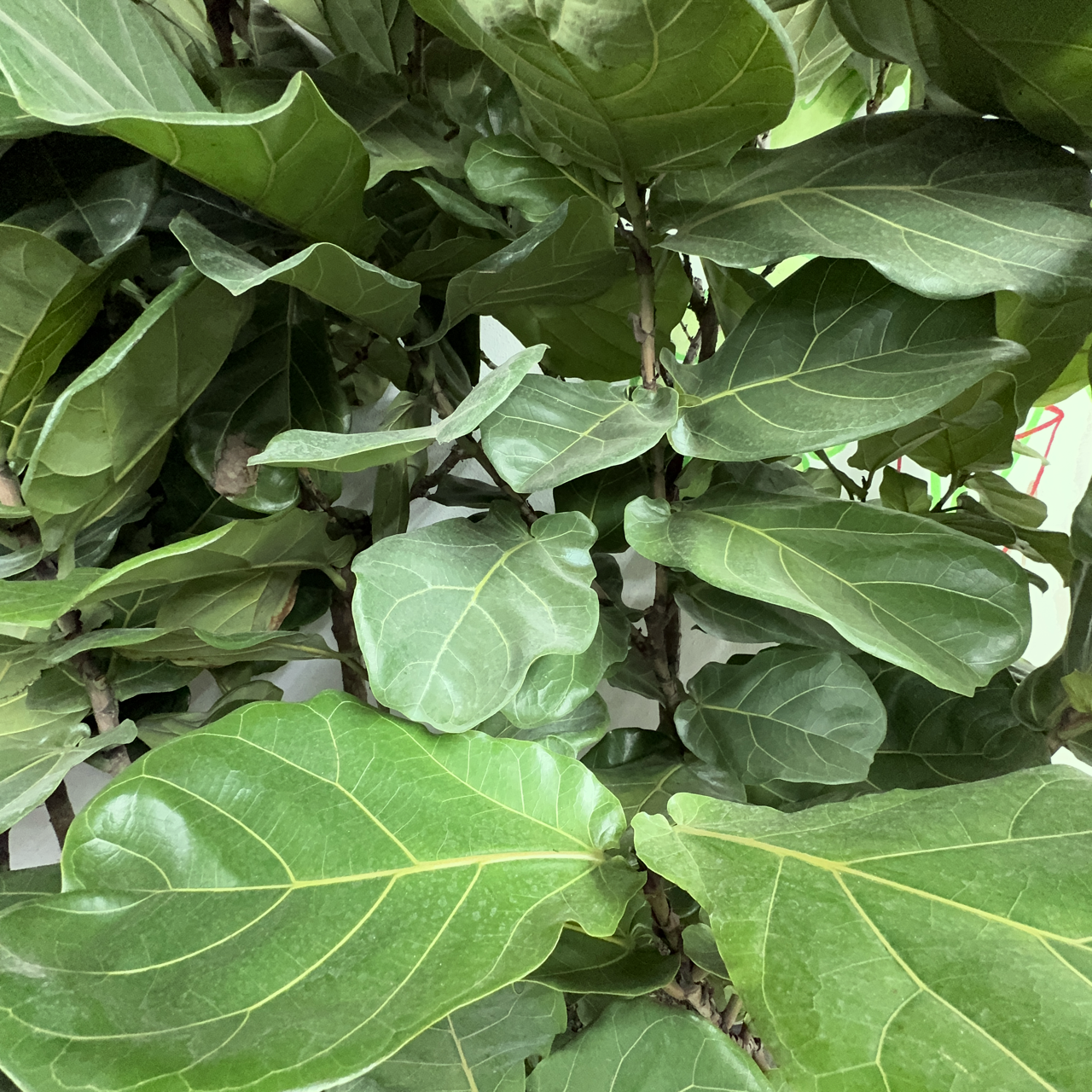}
            \end{subfigure}
            \vspace{2pt}
        \end{minipage}                     \\

        \centering \rotatebox{90}{\textbf{Leather}}   &
        \begin{minipage}{\linewidth}
            \centering
            \begin{subfigure}[b]{0.24\linewidth}
                \includegraphics[width=\linewidth, height=\linewidth]{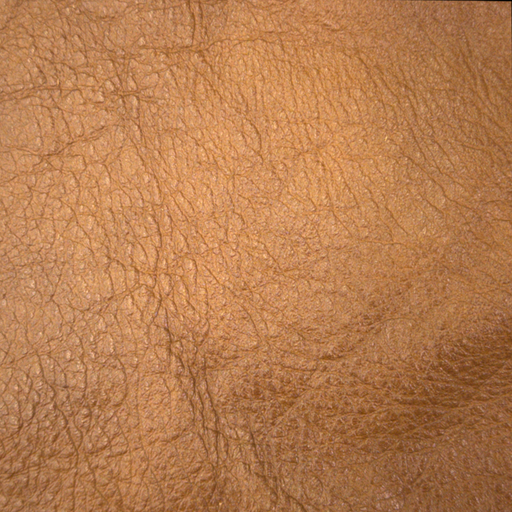}
            \end{subfigure}
            \hfill
            \begin{subfigure}[b]{0.24\linewidth}
                \includegraphics[width=\linewidth, height=\linewidth]{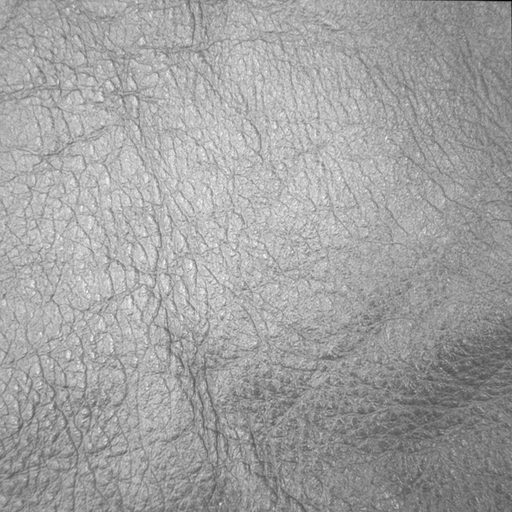}
            \end{subfigure}
            \hfill
            \begin{subfigure}[b]{0.24\linewidth}
                \includegraphics[width=\linewidth, height=\linewidth]{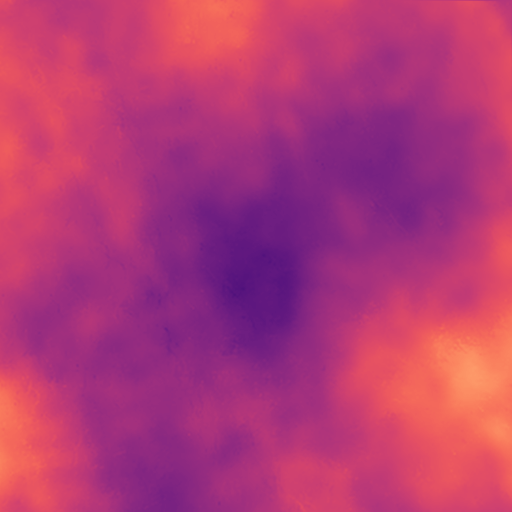}
            \end{subfigure}
            \hfill
            \begin{subfigure}[b]{0.24\linewidth}
                \includegraphics[width=\linewidth, height=\linewidth]{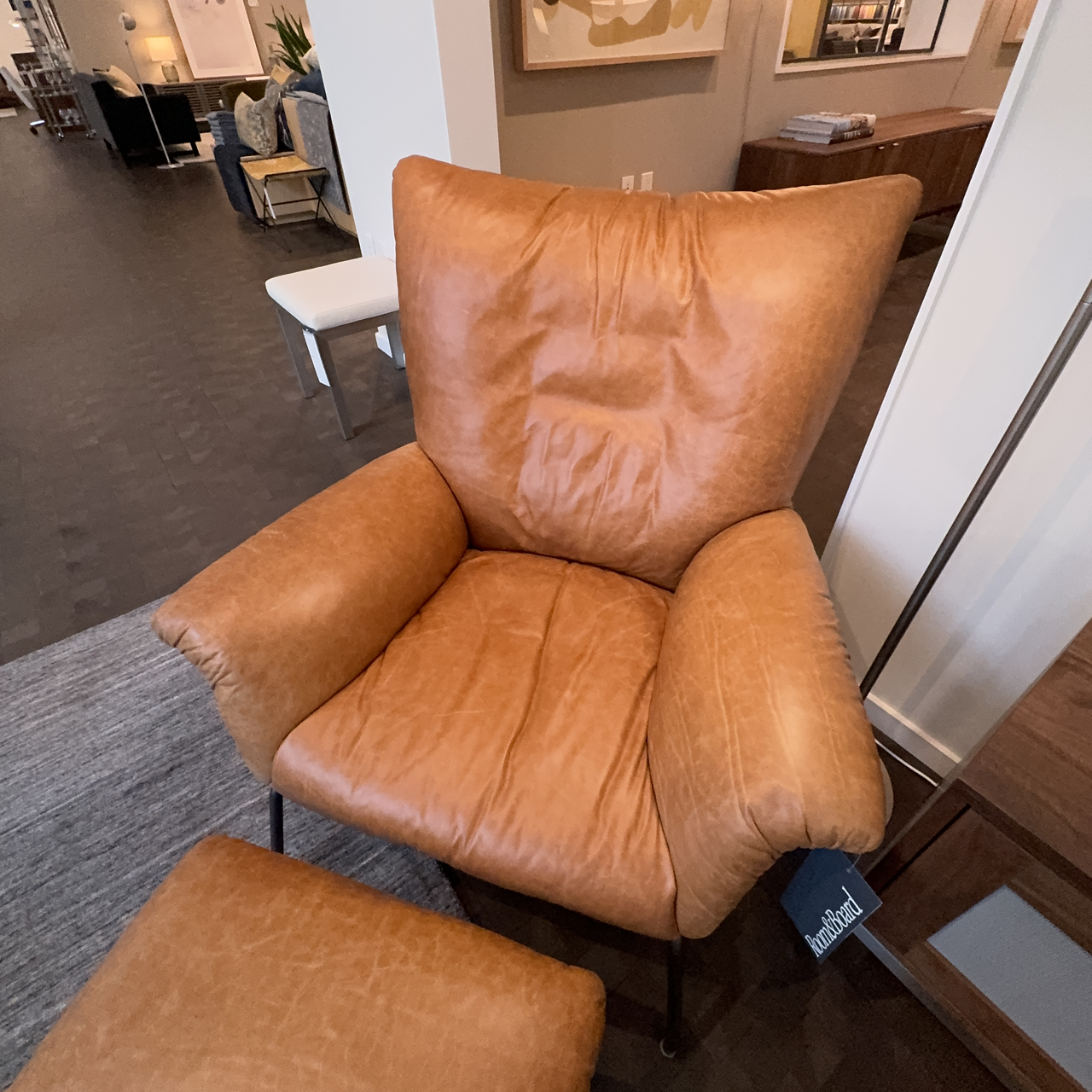}
            \end{subfigure}
            \vspace{2pt}
        \end{minipage}                     \\

        \centering \rotatebox{90}{\textbf{Tree Bark}} &
        \begin{minipage}{\linewidth}
            \centering
            \begin{subfigure}[b]{0.24\linewidth}
                \includegraphics[width=\linewidth, height=\linewidth]{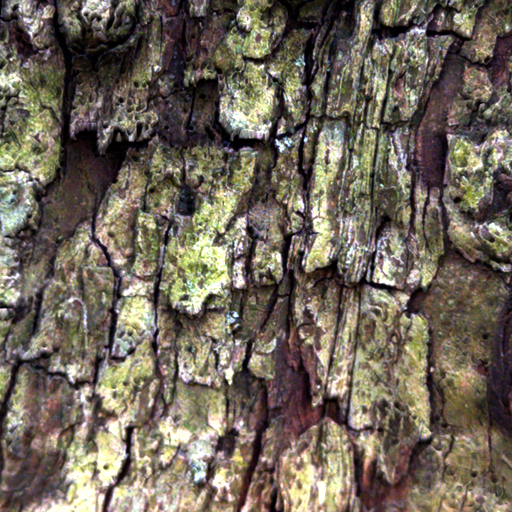}
            \end{subfigure}
            \hfill
            \begin{subfigure}[b]{0.24\linewidth}
                \includegraphics[width=\linewidth, height=\linewidth]{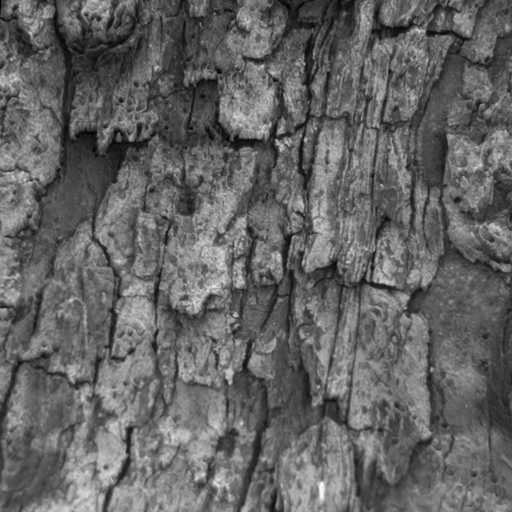}
            \end{subfigure}
            \hfill
            \begin{subfigure}[b]{0.24\linewidth}
                \includegraphics[width=\linewidth, height=\linewidth]{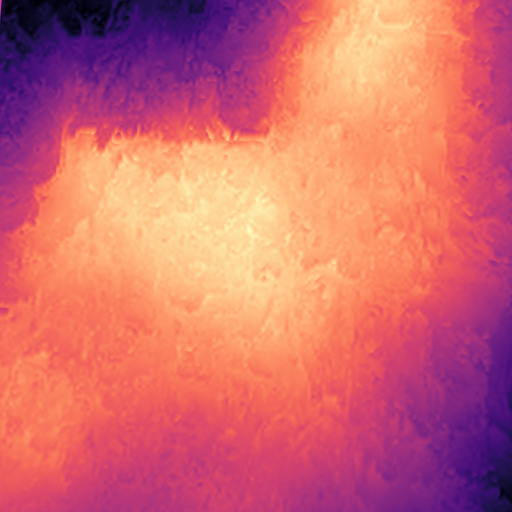}
            \end{subfigure}
            \hfill
            \begin{subfigure}[b]{0.24\linewidth}
                \includegraphics[width=\linewidth, height=\linewidth]{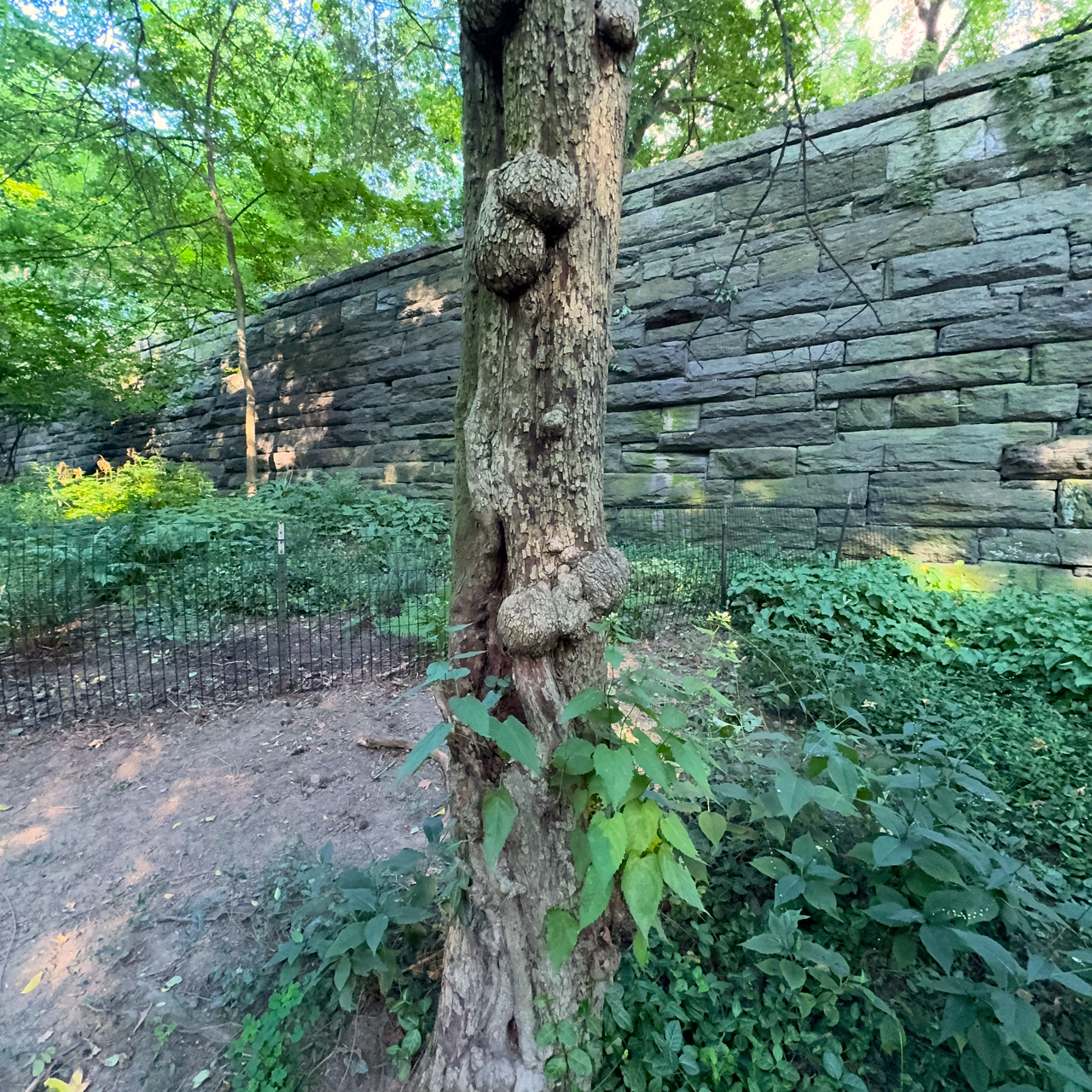}
            \end{subfigure}
            \vspace{2pt}
        \end{minipage}                     \\
    \end{tabular}

    \vspace{-2.5mm}
    \caption{
        \textbf{The \emph{Matador} Dataset.} (a-b)~Each sample includes a real-world image of a material taken at a high resolution and (c) its 3D structure (depth map). (d)~The surrounding context is also captured. The dataset comprises \(\sim\)7,200 samples across the 57 material categories of the proposed taxonomy (\cref{sec:taxonomy}).
    }\label{fig:dataset-mosaic}
    \vspace{-0.2in}
\end{figure}

\begin{figure*}
    \centering
    \tikzstyle{signal} = [rectangle, minimum width=1.5cm, minimum height=1cm, text centered, text width=2.3cm, draw=black, thick]
    \tikzstyle{noise} = [rectangle, minimum width=1.5cm, minimum height=1cm, text centered, text width=2.3cm, draw=black, dotted, thick]
    \tikzstyle{inout} = [rectangle, minimum width=1.5cm, minimum height=1cm, text centered, text width=2.3cm]
    \tikzstyle{arrow} = [thick,->,>=stealth]
    \tikzstyle{line} = [thick]

    \tikzset{
        right of/.append style={xshift=1.25cm},
        below of/.append style={yshift=-1.8cm},
    }

    \scalebox{0.62}{
        \large
        \begin{tikzpicture}
            \node (appearanceimage) [inner sep=3pt] {
                \setlength{\tabcolsep}{1pt}
                \begin{tabular}{m{3.75mm}m{27mm}}
                    \rotatebox{90}{Appearance} & \includegraphics[height=\linewidth, width=\linewidth,]{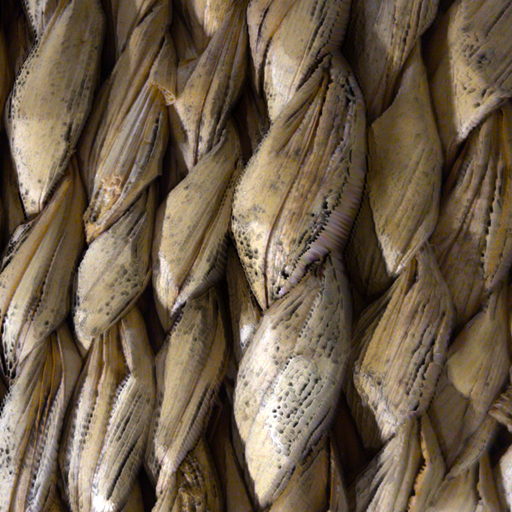}
                \end{tabular}
            };
            \node (depthimage) [inner sep=3pt, below of=appearanceimage] {
                \setlength{\tabcolsep}{1pt}
                \begin{tabular}{m{3.75mm}m{27mm}}
                    \rotatebox{90}{Depth Map} & \includegraphics[height=\linewidth, width=\linewidth,]{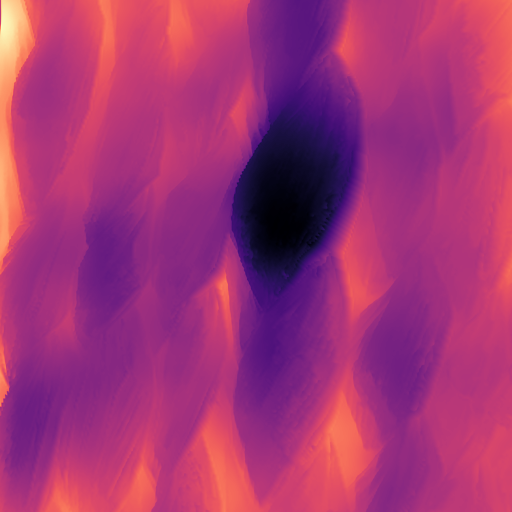}
                \end{tabular}
            };

            \node (mesh) [signal, right=1cm of appearanceimage] {3D Surface Mesh};

            \node (dotprod) [right of=mesh] {\scalebox{1.5}{$\odot$}};
            \node (phantomnode) [left=5mm of dotprod] {};
            \node (spatialaugment) [noise, below of=dotprod] {Spatial Transform};

            \node (radiance) [signal, right of=dotprod] {Scene Radiance};

            \node (conv1) [right of=radiance] {\scalebox{1.5}{$\circledast$}};
            \node (defocus) [noise, below of=conv1] {Lens Defocus};

            \node (opticalimage) [signal, right of=conv1] {Optical Image};

            \node (conv2) [right of=opticalimage] {\scalebox{1.5}{$\circledast$}};
            \node (pixelarea) [noise, below of=conv2, left=-5mm] {Pixel Active Area};

            \node (otimes) [right=5mm of conv2] {\scalebox{1.5}{$\otimes$}};
            \node (sampling) [noise, below of=otimes, right=-5mm] {Sampling (Pixel Pitch)};

            \node (discreteimage) [signal, right of=otimes] {Discrete Image};
            \node (oplus) [right of=discreteimage] {\scalebox{1.5}{$\oplus$}};
            \node (sensornoise) [noise, below of=oplus] {Photon \& Sensor Noise};

            \node (renderedimage) [inner sep=3pt, right=1cm of oplus] {
                \setlength{\tabcolsep}{2pt}
                \begin{tabular}{m{27mm}m{3.75mm}}
                    \includegraphics[height=\linewidth, width=\linewidth]{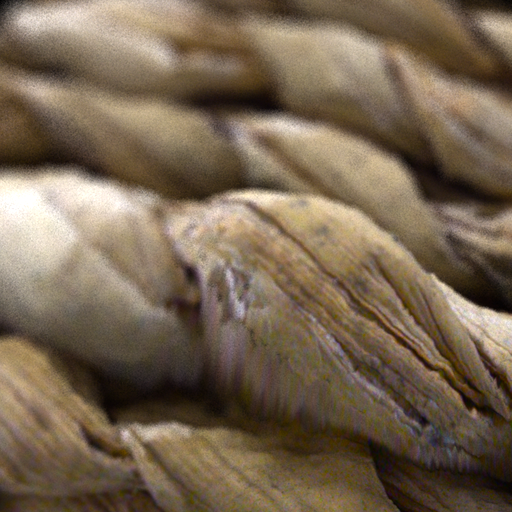} & \rotatebox[origin=rB]{-90}{Novel View}
                \end{tabular}
            };

            \draw [arrow] (appearanceimage) -- (mesh);
            \draw [arrow] (depthimage) -|- (mesh);

            \draw [arrow] (mesh) -- (dotprod);
            \draw [arrow] (spatialaugment) -- (dotprod);
            \draw [arrow] (dotprod) -- (radiance);

            \draw [arrow] (radiance) -- (conv1);
            \draw [arrow] (defocus) -- (conv1);

            \draw [arrow] (conv1) -- (opticalimage);
            \draw [arrow] (opticalimage) -- (conv2);
            \draw [arrow] (pixelarea) |-| (conv2);

            \draw [arrow] (conv2) -- (otimes);

            \draw [arrow] (sampling) |-| (otimes);
            \draw [arrow] (otimes) -- (discreteimage);

            \draw [arrow] (discreteimage) -- (oplus);
            \draw [arrow] (sensornoise) -- (oplus);

            \draw [arrow] (oplus) -- (renderedimage);

            \draw [decorate, decoration={brace, amplitude=10pt, mirror}]
            ([yshift=-3pt] spatialaugment.west|-spatialaugment.south) -- ([yshift=-3pt] spatialaugment.east|-spatialaugment.south)
            node[midway, below=10pt, align=center] {
                \textbf{(a)} Magnification \& Orientation
                \phantomsubcaption
                \label{fig:image-formation-model-augmentation}
            };

            \draw [decorate, decoration={brace, amplitude=10pt, mirror}]
            ([yshift=-3pt] defocus.west|-spatialaugment.south) -- ([yshift=-3pt] sensornoise.east|-spatialaugment.south)
            node[midway, below=10pt, align=center] {
                \textbf{(b)} Camera Settings
                \phantomsubcaption
                \label{fig:image-formation-model-rendering}
            };

        \end{tikzpicture}
    }
    \vspace{-0.25in}
    \caption{\textbf{Rendering Novel Views from a Real-World Sample.} From a captured material sample, we simulate its appearance under different magnifications, orientations, and camera settings. (a)~We first create a 3D mesh and texture map it with the appearance image. We then apply spatial transformations to the mesh to change its pose. (b)~The optical image of a novel view (including depth of field effects) is obtained by raytracing. It is then blurred to account for pixel area, and the result is sampled to produce the discrete image. Finally, noise is added, resulting in the novel view. By varying the parameters in this process, we render numerous novel views for each real-world sample.}\label{fig:image-formation-model}
    \vspace{-0.15in}
\end{figure*}
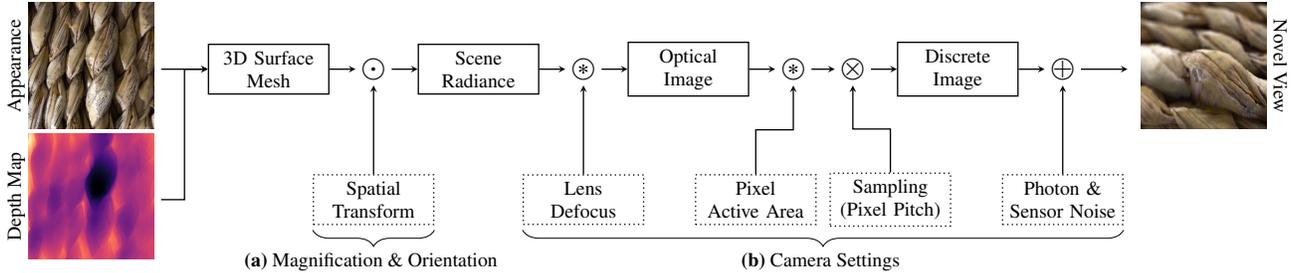

To populate our taxonomy, we collected a dataset, named \emph{Matador}, that consists of over 7,200 samples across the 57 different materials in our taxonomy. \cref{fig:dataset-mosaic} shows examples of samples in the dataset. For each sample, we capture a close-up, high-resolution image of the local appearance (color in \cref{fig:dataset-mosaic-color}, grayscale in \cref{fig:dataset-mosaic-grayscale}) and a registered lidar scan of the material's 3D surface structure (\cref{fig:dataset-mosaic-depth}). The 3D structure was captured to enable us to generate additional views of the material, corresponding to different magnifications, orientations, and camera settings. A wide-angle image of the sample in its surrounding context is also captured (\cref{fig:dataset-mosaic-context}). \emph{Matador} is unique in three respects: (a) the diversity of materials it includes, (b) the hierarchical labels assigned to samples, and (c) the use of 3D structure to generate novel views. It is available on the \href{https://cave.cs.columbia.edu/repository/Matador}{\emph{Matador} webpage}.

\textbf{Data Collection.}
To collect the dataset, we developed an iOS application. A smartphone allows for increased mobility when capturing in-the-wild samples, meaning the user can image the sample at the camera's minimum focus distance to maximize spatial resolution. We chose the iPhone 15 Pro Max as our platform for the quality of its cameras and lidar.\footnote{Among tested smartphones, the iPhone 15 Pro Max has the only lidar capable of capturing depth maps at close distances (around the camera's minimum focus distance of \(\sim\)15~cm). We use lidar since stereo matching can be unreliable for weakly textured materials. Additionally, the iPhone 15 Pro Max has superior optical performance with a measured MTF50 of 0.240~cycles/pixel compared to, \eg, the Pixel 8 Pro's 0.158~cycles/pixel.} For each sample, the local appearance and 3D structure are captured using the wide-angle camera (12MP, 74\textdegree~FOV) and lidar (100~\unit{points \per degree \squared}, 74\textdegree~FOV). Subsequently, an additional context image is taken from a more distant viewpoint using the ultrawide-angle camera (12MP, 104\textdegree~FOV). Both images (appearance and context) are 12-bit Bayer raw. For cases where one may wish to register the local and context images, we also record the phone's motion during the time between the two captures using the phone's inertial measurement unit. Since the data collection is done in the wild, lighting and camera viewpoint vary across samples. We have publicly released the iOS application, enabling \emph{Matador} to grow in size and scope over time.

\textbf{Rendering Novel Views.}
Using the 3D structure of the captured sample, we can simulate images under different magnifications, orientations, and camera settings. This process is outlined in \cref{fig:image-formation-model}. We begin by creating a 3D mesh from the depth map and texture-mapping it with the surface radiance values measured in the local appearance image. Spatial transformations are then applied to the mesh to change the pose of the original sample. For the purpose of view generation, we assume the sample to be Lambertian
and do not alter illumination (each \emph{Matador} class already captures a wide variety of point and extended sources). A high-resolution (optical) image for a novel viewpoint is then generated by raytracing the mesh, using a thin-lens model to include depth-of-field (defocus) effects~\cite{jakobDRJITJustintimeCompiler2022}. Raytracing implicitly accounts for occlusions and perspective effects, which are both important in the case of significant surface depth variations. To produce the discrete image, we first blur the optical image with a box filter whose support is equal to the pixel active area. The result is then sampled by a pulse train with period equal to the pixel pitch. Finally, we add photon and sensor noise to produce the novel view.

\begin{figure}
    \centering
    \footnotesize

    \setlength{\tabcolsep}{1pt}
    \begin{tabular}{m{0.99\columnwidth}}
        \begin{minipage}{\linewidth}
            \centering
            \begin{subfigure}[b]{0.24\linewidth}
                \setlength{\fboxsep}{-2pt}
                \setlength{\fboxrule}{2pt}
                \fbox{\includegraphics[width=\linewidth, height=\linewidth]{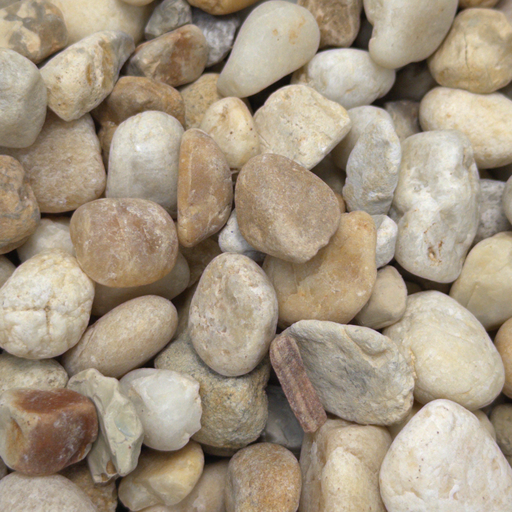}}
            \end{subfigure}
            \hfill
            \begin{subfigure}[b]{0.24\linewidth}
                \includegraphics[width=\linewidth, height=\linewidth]{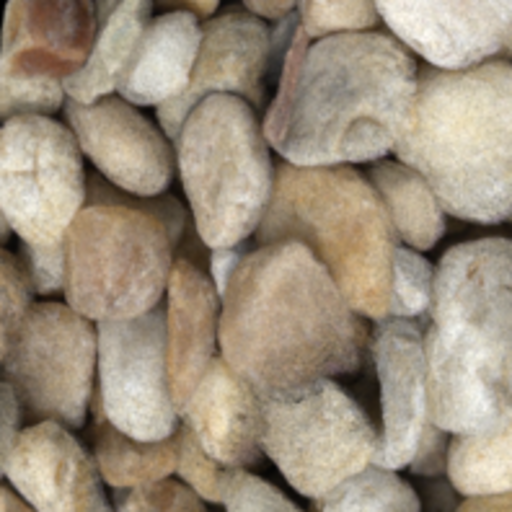}
            \end{subfigure}
            \hfill
            \begin{subfigure}[b]{0.24\linewidth}
                \includegraphics[width=\linewidth, height=\linewidth]{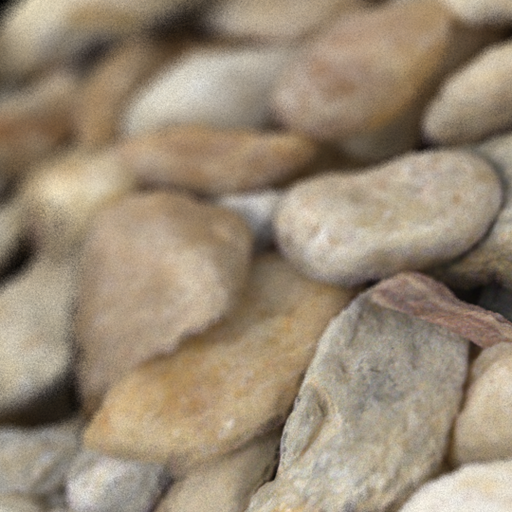}
            \end{subfigure}
            \hfill
            \begin{subfigure}[b]{0.24\linewidth}
                \includegraphics[width=\linewidth, height=\linewidth]{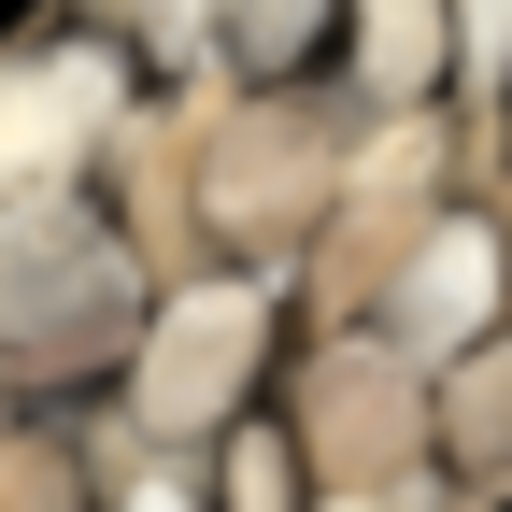}
            \end{subfigure}
            \vspace{\dimexpr\tabcolsep+1pt\relax}
        \end{minipage} \\
    \end{tabular}

    \vspace{-2.5mm}
    \caption{
        \textbf{Rendered Novel Views for Gravel.} Examples of the many novel views rendered from a single real-world sample of gravel (left) using the process described in \cref{sec:dataset}.
    }\label{fig:novel-views}
    \vspace{-0.2in}
\end{figure}

By altering the pose of the sample, defocus of the lens, pixel active area, pixel pitch, and image noise level, we can simulate how an arbitrary camera would image the sample at any magnification and orientation. Many novel views are rendered for each sample (see \cref{fig:novel-views}). This process is applied to all \(\sim\)7,200 raw samples in the \emph{Matador} dataset to obtain a larger and more diverse set of material images to supplement model training, which we show improves generalization to real-world settings (\cref{sec:method}).

\textbf{Generalization Test Set.}
To evaluate the impact novel views have on recognition performance in the wild, we construct an additional out-of-distribution (OOD) test set. Given a material sample in \emph{Matador}, we begin with the region of interest (ROI) defined in the appearance image. Using the depth map and motion between appearance and context captures, we map the ROI onto the context image and extract a patch of similar scale. Accelerometer noise means this patch does not exactly align with the ROI seen in the appearance image, and instead captures a different area on that instance of material. In addition, as the context image has a different viewpoint, image sensor, and field of view, the resultant patches comprise an OOD test set. See \cref{sec:supp-generalization-test-set} of the supplemental material for further details on the construction of this test set, as well as example images.

\section{Recognizing Materials}\label{sec:method}
We now describe a method that leverages our proposed taxonomy to recognize materials from their local appearance. Consider the materials ``iron'' and ``brass'' in \cref{fig:taxonomy-tree}; they are distinct classes but share a common ancestor, ``metal'', which is characterized by a range of mechanical properties. They are arranged in the taxonomy according to their mechanical properties, and we speculate that materials with close taxonomic proximity may share visual traits. While not guaranteed, should such similarities exist, our aim is to exploit them for recognition. In addition, in cases where we are unable to recognize a local appearance as ``brass'', it would still be useful for a downstream application to know that it is a metal. For this reason, we design our recognition method to account for the structure of our taxonomy. To this end, we use a graph neural network (GNN) to encode the structure of the taxonomy and thereby constrain how materials share features in latent space (based on their taxonomic proximity). While training this model, our objective is to predict an image's full taxonomic classification, \ie, the class at each level of the taxonomy. We refer to this approach as \emph{hierarchical material recognition}.

To achieve hierarchical recognition, the proposed taxonomy is first translated into a directed graph $\mathcal{G}=(\mathcal{V},\mathcal{E})$, where $\mathcal{V}$ is the set of nodes in the taxonomy and $\mathcal{E}$ is the set of edges that represent parent-child relationships. We then treat recognition as a node classification task. Given a set of images $\mathcal{T}$, the label for each image $x$ is the hierarchical path of the material in the taxonomy $\ell \subseteq \mathcal{V}$. For example, the label of an image of steel would be \{solid, abiotic, metal, ferrous, steel\}. The set of all images whose label set~$\ell$ contains the node $v_i$ is then $\mathcal{T}_i = \{ x_j \in \mathcal{T} \, | \, v_i \in \ell_j \}$. We initialize a learnable vector for node $v_i$ in our graph model with the average image features over $\mathcal{T}_i$, produced by an encoder $\phi$ such as a convolutional neural network. Explicitly, the initial feature vector of $v_i$ is $\mathbf{h}_i^{0} = \frac{1}{||\mathcal{T}_i||} \sum_{x_j \in \mathcal{T}_i} \phi(x_j)$. To classify an image $x$, a global node $\mathbf{h}_g = \phi(x)$ is inserted into the graph with outgoing edges to every other node.

For our construction, we use a graph attention network (GAT)~\cite{velickovicGraphAttentionNetworks2018}. The message passing update for a node $v_i$ at level $k$ of the model is therefore defined by the relation:
\begin{align}
  \label{eq:gat-update}
  \mathbf{h}_i^{k+1} & = \psi_a \left( \mathbf{h}_i^{k}, \oplus_{j\in \mathcal{N}_i} \left( \alpha_{ij}^k \, \psi_b \left( \mathbf{h}_i^{k}, \mathbf{h}_j^{k}\right) \right) \right),
\end{align}
where $\psi_a$ and $\psi_b$ are multi-layer perceptrons. The local neighborhood $\mathcal{N}_i$ of node $v_i$ is fixed by the adjacency matrix of $\mathcal{G}$, and $\oplus$ is a permutation-invariant aggregation function (summation in our case). The edge attention mechanism $\alpha_{ij}^{k}$ denotes the connective strength between node $v_i$ and its neighbor $v_j$, and this is used to weigh the sharing of visual traits amongst taxonomy nodes (\eg, from individual child nodes to their parent). Since \cref{eq:gat-update} only operates on the local neighborhood of each node, we cascade up to $D$ layers to match the diameter of the taxonomy and propagate updates throughout the model. Residual connections are also added between layers to mitigate oversmoothing gradients.

We jointly train $\phi$ and the GAT classifier ($\alpha, \psi_a, \psi_b$) end-to-end, where $\phi$ is a ResNet50~\cite{heDeepResidualLearning2016} initialized by pretraining on the IG-1B~\cite{mahajanExploringLimitsWeakly2018} and ImageNet1k~\cite{dengImageNetLargescaleHierarchical2009} datasets. As our goal is multi-label hierarchical classification, we use binary cross-entropy (BCE) as our loss on the label set $\ell=[\ell^0, \ell^1, \ldots, \ell^{D-1}]$. We also use cross-entropy (CE) as our loss for each hierarchy level. The full loss is a greedy combination of both:~$\max ( \text{BCE}(\hat{\ell}, \ell),\, \frac{1}{D}\sum_{d=0}^{D-1} \text{CE} ( \hat{\ell}^d, \ell^d ))$, where the first term encourages learning complete hierarchical paths, and the second term encourages correct predictions within each level of the hierarchy.

\subsection{Flat Classification}
To validate our model, we first compare its performance with recent methods on standard material recognition datasets. These local appearance datasets include KTH-TIPS2-b~\cite{caputoClassspecificMaterialCategorisation2005}, FMD~\cite{sharanRecognizingMaterialsUsing2013}, GTOS~\cite{xueDifferentialAngularImaging2017}, and GTOS-Mobile~\cite{xueDeepTextureManifold2018}. KTH-TIPS2-b extends CUReT~\cite{danaReflectanceTextureRealworld1999} and is a BTF dataset where materials are imaged in a controlled setting with different magnifications, orientations, and illuminations. FMD includes a similar number of materials, but the images are instead aggregated from online sources. GTOS is larger in scope, comprising radiometric images of ground terrain captured using specialized hardware. GTOS-Mobile extends GTOS by adding smartphone images for a subset of the original GTOS materials. Further dataset details can be found in the supplemental material (\cref{tab:existing-datasets}).

To evaluate the performance of our model, categories from the above datasets are mapped into our taxonomy, renaming and inserting leaf nodes as necessary. This process does not alter the high-level structure of the taxonomy. We measure test accuracy on multi-class flat classification after training our model with multi-label hierarchical learning. The top-1 accuracy is reported in \cref{tab:top-1-accuracy}, where all models use the same ResNet50 backbone. We can see our hierarchical, graph-based classifier results in a significant performance increase with respect to existing methods.

\begin{table}
  \centering
  \footnotesize
  \begin{threeparttable}
    \caption{
      \textbf{Efficacy of Graph Representation Learning.} We compare our hierarchical model with recent material recognition methods. In all cases, including ours, a ResNet50 feature extractor was used. The best overall results are highlighted in bold, and the second best are underlined. Leveraging the structure of our taxonomy through graph representation learning yields significant improvements in recognition performance.
      \vspace{-0.1in}
    }\label{tab:top-1-accuracy}
    \setlength{\tabcolsep}{6pt}
    \begin{tabular}{lllll}
      \toprule
                                                                  & \multicolumn{4}{c}{Top-1 Accuracy \(\uparrow\)}                                                                                                                             \\
      \cmidrule(l{\tabcolsep}r{\tabcolsep}){2-5}
      Method                                                      & KTH-2-b                                         & FMD                                      & GTOS                                        & GTOS-M                           \\
      \midrule
      DeepTEN~\cite{zhangDeepTENTexture2017}                      & \(82.0 {\scriptstyle \pm 3.3}\)                 & \(80.2 {\scriptstyle \pm 0.9}\)          & \(84.5 {\scriptstyle \pm 2.9}\)             & ---\textsuperscript{\textdagger} \\
      MAPNet~\cite{zhaiDeepMultipleAttributePerceivedNetwork2019} & \(84.5 {\scriptstyle \pm 1.3}\)                 & \(85.7 {\scriptstyle \pm 0.7}\)          & \(84.7 {\scriptstyle \pm 2.2}\)             & \(86.6\)                         \\
      DSRNet~\cite{zhaiDeepStructureRevealedNetwork2020}          & \(85.9 {\scriptstyle \pm 1.3}\)                 & \(86.0 {\scriptstyle \pm 0.8}\)          & \(85.3 {\scriptstyle \pm 2.0}\)             & \underline{\(87.0\)}             \\
      FENet~\cite{xuEncodingSpatialDistribution2021}              & \(86.6 {\scriptstyle \pm 0.1}\)                 & \(82.3 {\scriptstyle \pm 0.3}\)          & \(83.1 {\scriptstyle \pm 0.2}\)             & \(85.1\)                         \\
      CLASSNet~\cite{chenDeepTextureRecognition2021}              & \(87.7 {\scriptstyle \pm 1.3}\)                 & \(86.2 {\scriptstyle \pm 0.9}\)          & \underline{\(85.6 {\scriptstyle \pm 2.2}\)} & \(85.7\)                         \\
      DFAEN~\cite{yangDFAENDoubleorderKnowledge2022}              & \(86.3\)                                        & \(86.9\)                                 & ---\textsuperscript{\textdagger}            & \(86.9\)                         \\
      RADAM~\cite{scabiniRADAMTextureRecognition2023}             & \(88.5 {\scriptstyle \pm 3.2}\)                 & \(85.3 {\scriptstyle \pm 0.4}\)          & \(81.8 {\scriptstyle \pm 1.1}\)             & \(81.0\)                         \\
      FRP~\cite{florindoFractalPoolingNew2024}                    & \underline{\(90.7\)}                            & \underline{\(88.8\)}                     & ---\textsuperscript{\textdagger}            & ---\textsuperscript{\textdagger} \\
      \midrule
      Ours                                                        & \(\mathbf{93.5 {\scriptstyle \pm 4.0}}\)        & \(\mathbf{96.1 {\scriptstyle \pm 0.6}}\) & \(\mathbf{87.9 {\scriptstyle \pm 2.1}}\)    & \(\mathbf{92.2}\)                \\
      \bottomrule
    \end{tabular}
    \begin{tablenotes}
      \scriptsize
      \item[\textdagger] Value not reported in the original work.
    \end{tablenotes}
  \end{threeparttable}
  \vspace{-0.2in}
\end{table}

Next, we evaluate the performance of our model on our \emph{Matador} dataset. For this evaluation, we emphasize performance on the materials in \emph{Matador} that we deem to have sufficient texture to be identifiable using local appearance alone. We create a subset of \emph{Matador} where materials such as glass, plastics, and paint are omitted as they are flat in appearance. In addition, some material classes are consolidated into a single class. For instance, all metals have similar appearances (finishes) except for changes in hue, and hence are combined into a single category. With these omissions and consolidations, we arrived at a dataset containing 37 material classes with \(\sim\)6,600 samples. We refer to this slice of \emph{Matador} as \emph{Matador-C1}, and details of its construction are given in the supplemental material (\cref{sec:sup-datasets}).

\begin{figure}
  \centering
  \begin{minipage}[b]{\linewidth}
    \centering
    \footnotesize
    \begin{threeparttable}
      \subcaption{Top-1 accuracy, ablating novel views during training of our models.}\label{fig:matador-accuracy:benchmarks}
      \setlength{\tabcolsep}{2.35pt}
      \begin{tabular}{llllll}
        \toprule
                                                             &                                        & \multicolumn{4}{c}{Top-1 Accuracy \(\uparrow\)}                                                                                                                                                                                                  \\
        \cmidrule(l{\tabcolsep}r{\tabcolsep}){3-6}
        Method                                               & Params                                 & Matador                                         & Matador-C1           & \multicolumn{2}{l}{Out-of-Distrib.\textsuperscript{\textdagger}}                                                                                                        \\
        \midrule
        \multicolumn{6}{l}{\emph{Vision Foundation Models (Zero-Shot)}} \vspace{1pt}                                                                                                                                                                                                                                                                     \\
        \, CLIP~\cite{radfordLearningTransferableVisual2021} & 151~M                                  & \(24.8\)                                        & \(40.0\)             & \(32.3\)                                                                                                                                                                \\
        \, GPT-4.1~\cite{IntroducingGPT41API}                & 1.76~T\textsuperscript{\textdaggerdbl} & \(51.4\)                                        & \(65.9\)             & \(53.4\)                                                                                                                                                                \\
        \midrule
        \multicolumn{6}{l}{\emph{Material Recognition Models (ResNet50 Backbone, Finetuned)}} \vspace{1pt}                                                                                                                                                                                                                                               \\
        \, DeepTEN~\cite{zhangDeepTENTexture2017}            & 24~M                                   & \(79.2\)                                        & \(88.8\)             & \(61.5\)                                                                                                                                                                \\
        \, DEPNet~\cite{xueDeepTextureManifold2018}          & 25~M                                   & \(82.7\)                                        & \(87.6\)             & \(76.1\)                                                                                                                                                                \\
        \, FRP~\cite{florindoFractalPoolingNew2024}          & 28~M                                   & \(74.8\)                                        & \(89.4\)             & \(71.0\)                                                                                                                                                                \\
        \, MSLac~\cite{mohanLacunarityPoolingLayers2024}     & 24~M                                   & \(82.6\)                                        & \(88.5\)             & \(75.4\)                                                                                                                                                                \\
        \midrule
        \multicolumn{6}{l}{\emph{Modern Image Recognition Models (Finetuned)}} \vspace{1pt}                                                                                                                                                                                                                                                              \\
        \, ConvNext-V2~\cite{wooConvNeXtV2CoDesigning2023}   & 28~M                                   & \(83.1\)                                        & \(89.7\)             & \(81.9\)                                                                                                                                                                \\
        \, EVA-02~\cite{fangEVA02VisualRepresentation2024}   & 22~M                                   & \(85.8\)                                        & \(90.1\)             & \(82.7\)                                                                                                                                                                \\
        \midrule
        \multicolumn{6}{l}{\emph{\textbf{Hierarchical Material Recognition Models (Various Backbones)}}} \vspace{1pt}                                                                                                                                                                                                                                    \\
        \, Ours {\scriptsize(ResNet50)}                      & 28~M                                   & \(85.8\)                                        & \(94.1\)             & \(82.9\) {(+\(2.8\))}                                            & \hspace{-0.185cm}\footnotesize\rdelim\}{3}{*}[\tiny\parbox{0.5cm}{\centering Gain from Novel Views}] \\
        \, Ours {\scriptsize(ConvNext-V2)}                   & 31~M                                   & \underline{\(85.9\)}                            & \underline{\(94.2\)} & \underline{\(86.8\)} {(+\(2.4\))}                                                                                                                                       \\
        \, Ours {\scriptsize(EVA-02)}                        & 24~M                                   & \(\mathbf{88.3}\)                               & \(\mathbf{94.7}\)    & \(\mathbf{87.0}\) {(+\(4.9\))}                                                                                                                                          \\
        \bottomrule
      \end{tabular}
      \begin{tablenotes}
        \scriptsize
        \item[\textdagger] Train on appearance images, evaluate on patches of context images.
        \item[\textdaggerdbl] Unreported for GPT-4.1, value reported is for its predecessor GPT-4.
      \end{tablenotes}
    \end{threeparttable}
    \vfill
  \end{minipage}
  \begin{minipage}[b]{\linewidth}
    \centering
    \vspace{0.05in}
    \includegraphics[width=\linewidth]{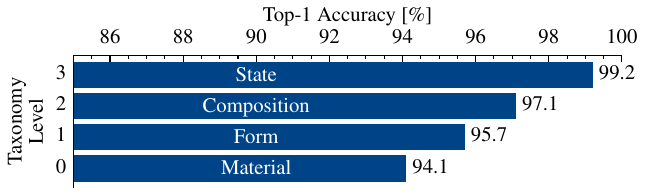}
    \subcaption{``Ours (ResNet50)'' top-1 accuracy on \emph{Matador-C1} by taxonomy level.}\label{fig:matador-accuracy:levels}
    \vspace{-0.1in}
  \end{minipage}
  \vspace{-0.1in}
  \caption{
    \textbf{Performance on \emph{Matador}.} (a)~Comparison with existing methods. The best overall results are highlighted in bold, and the second best are underlined. Our model demonstrates state-of-the-art recognition, and supplementing training with rendered novel views enhances performance on out-of-distribution (OOD) data (see \cref{sec:dataset}). (b)~Hierarchical classification accuracy. Note that accuracy increases with the taxonomy level, suggesting that exploited visual similarities at lower levels improve recognition at higher levels, and that an image may be correctly classified at a higher level (\eg, form) even when its material class is uncertain.
  }\label{fig:matador-accuracy}
  \vspace{-0.2in}
\end{figure}

The performance of our method and existing methods on the full \emph{Matador} dataset and the consolidated \emph{Matador-C1} dataset is summarized in \cref{fig:matador-accuracy:benchmarks}. All finetuned models are trained with the same optimizers (Muon~\cite{jordanMuonOptimizerHidden2025} and AdamW~\cite{loshchilovFixingWeightDecay2018}). Compared to previous methods, we again achieve state-of-the-art performance. We additionally show how supplementing the training data with rendered novel views considerably improves generalization to OOD imaging conditions (up to 4.9\%). We refer the reader to \cref{sec:sup-evaluation} of the supplemental material for class-specific accuracies of our model, its performance on grayscale images, and competing material recognition model performance when only finetuning the classifier head instead of the entire model.

\subsection{Hierarchical Classification}\label{sec:hierarchical-classification}
Though we use hierarchical inference during training of our model, to this point, we have only reported performance for flat classification of the taxonomy leaves. We now examine the capability of our model for multi-label hierarchical classification. Standard methods for hierarchical classification use a deterministic~\cite{shottonSemanticTextonForests2008} or probabilistic~\cite{gaoDeepHierarchicalClassification2020} walk down the levels of the tree. Using these techniques, there is a risk of outputting invalid label combinations unless care is taken to enforce the hierarchy structure (\eg, through a specialized loss function or inference routine). In our case, we encode the hierarchy structure in the GNN and also use a hierarchical loss function. For this evaluation, however, we do not enforce hierarchical consistency in the inference routine (which we do later in \cref{sec:discussion}). In \cref{fig:matador-accuracy:levels}, we present classification accuracy on \emph{Matador-C1} for each level of the taxonomy. As expected, we see higher performance at higher levels of the taxonomy where the number of classes is fewer. This result suggests that using a GNN to encode the taxonomy structure effectively guides hierarchical learning by exploiting any shared visual traits amongst taxonomy nodes. It also implies that when we are unable to correctly recognize an image at the finest level (material), we may still be able to recognize it at a coarser one (\eg, form).


\begin{figure*}
    \centering
    \includegraphics[width=0.98\linewidth]{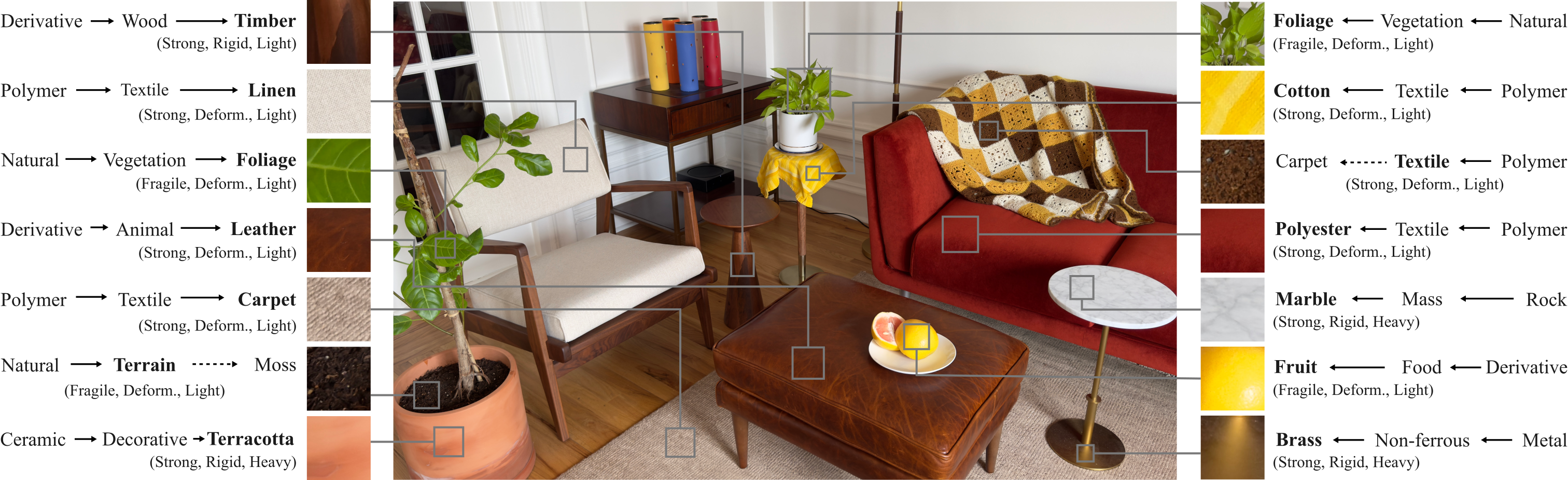}
    \vspace{-1mm}
    \caption{
        \textbf{Probing Materials in a Scene.} Given an image, we can probe the material at a point to determine its taxonomic classification. For each probed point, we automatically find the window size that maximizes classification confidence. Each classification is hierarchical, and the finest level that exceeds a confidence threshold is shown in bold. We can associate the predicted material with its known mechanical properties to provide information that could be useful, \eg, for robot manipulation. These properties (\eg, ``Strong, Rigid, Light'') are listed below the finest classification achieved by our model. Strong v.\ Fragile is determined by a tensile strength threshold, Rigid v.\ Deformable by a Young's modulus threshold, and Heavy v.\ Light using a density threshold.\protect\footnotemark~~Note that when we are unable to identify the exact material (\eg, the wool blanket misclassified as carpet), we still achieve correct classification (``textile'') at a coarser level in the taxonomy.
    }\label{fig:probing}
    \vspace{-0.15in}
\end{figure*}

\section{Implications and Extensions}\label{sec:discussion}
We now show how our proposed model may be used in practice and discuss potential extensions of our work.

\textbf{Probing Materials in a Scene.}
In many vision tasks, such as navigation, it is desirable to know the material at an arbitrary point in a scene. That is, we would like to query a pixel in an image to determine its material. As noted in \cref{sec:dataset}, materials may appear very different at different scales. Therefore, what is a good strategy for selecting an appropriate region around a probed image point to use for classification? If depth is available, one could fix the window size using metric units. Here, as in \cref{sec:method}, we focus on the more common case where depth is not available during inference. Given a probed image pixel, we use windows of increasing size from 64\(\times\)64~px\textsuperscript{2} to 1024\(\times\)1024~px\textsuperscript{2}. These windows are passed through our model using Monte Carlo dropout~\cite{kendallWhatUncertaintiesWe2017,galDropoutBayesianApproximation2016} to build a distribution of the predictions. This distribution and its entropy then guide a best-first search of the taxonomy tree to obtain a hierarchically consistent classification for the chosen image point.

\cref{fig:probing} shows the above probing approach applied to several points in an imaged scene. In each case, the window that produced the highest confidence classification is displayed. Note that all probed points in this scene are at significantly greater distances (lower magnifications) than the training samples in \emph{Matador}. Yet, we are able to correctly classify most of the points due to the rendered novel views used during training. In cases where we are unable to correctly identify the low-level material, we can still recognize it at a higher level of the taxonomy. Two examples of this are the wool blanket on the couch that was classified as carpet and the soil in the planter that was classified as moss. A natural extension of this method is to apply probing to all pixels in the image and then use neighboring pixels to inhibit or reinforce the classification of each pixel. The result would be a material-based segmentation of the entire scene.

\begin{figure}
    \centering
    \includegraphics[width=0.99\linewidth]{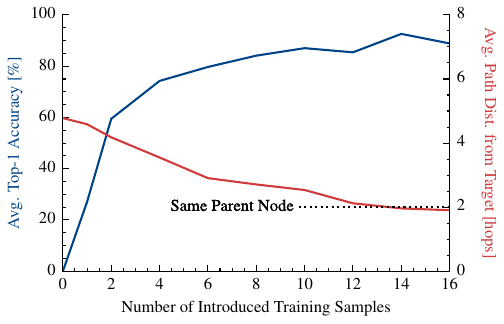}
    \vspace{-2.5mm}
    \caption{
        \textbf{Few-Shot Learning of New Materials.} (\textcolor{plotblue}{Blue Plot})~Average performance for classes not seen during pretraining, as a function of the number of training samples reintroduced during finetuning. The plot shows that novel classes can be learned with a small number of samples, achieving \(\sim\)\(90\%\) accuracy with just 16 samples. (\textcolor{plotred}{Red Plot})~With the same number of samples, misclassifications are, on average, localized to siblings at the lowest level of the taxonomy (\ie, ``Materials'' with the same ``Form'').
    }\label{fig:nshot}
    \vspace{-0.2in}
\end{figure}

\textbf{What about Unseen Materials?}
We now discuss an important extension to material recognition: few-shot learning~\cite{fei-feiOneshotLearningObject2006,satorrasFewShotLearningGraph2018,guoNeuralGraphMatching2018,liuPrototypePropagationNetworks2019} of novel categories. Such a capability is desirable for an intelligent system to infer the mechanical properties of previously unseen materials. We evaluate our hierarchical model in this setting by training copies on modified versions of the \emph{Matador-C1} dataset, each having a single material category held out. In \cref{fig:nshot}, the blue plot shows the average accuracy of our model on the held-out class as a function of the number of held-out samples reintroduced into training. Note that novel material classes are learned rapidly. The red plot shows the average path distance, \ie, the number of edges (hops) between the predicted node and the correct node in the taxonomy. We observe that, with as few as 16 samples of a novel class, we achieve an average accuracy of \(\sim\)\(90\%\) and an average path distance of less than \(2\) hops. The latter suggests that our model, using a small number of samples, is able to learn the hierarchical label of a material in our taxonomy that has not been seen before. Even when the model is unable to correctly recognize the material, it is able to identify it correctly at one level higher in the taxonomy. This potential for rapid adaptation could enable intelligent systems to learn novel materials added to our taxonomy (\eg, liquids and gases) with little expense.

\section{Summary}
In this work, we have introduced a taxonomy of materials suited for visual perception, a diverse dataset (\emph{Matador}) of material images, a rendering-based training augmentation that improves OOD model generalization, and a graph representation method that leverages the taxonomy for visual recognition. On existing datasets and \emph{Matador}, we have shown the efficacy of hierarchical learning using graph attention. We demonstrated the ability of this model to estimate the taxonomic class of a material, and to quickly adapt to unseen materials through few-shot learning. We believe that hierarchical material recognition can help intelligent systems perform more sophisticated tasks and, in addition, achieve higher levels of safety and reliability.

\footnotetext{Thresholds used are 1 \unit{\MPa}, 12 \unit{\GPa} and 1,600 \unit{\kilogram\per\meter\cubed}, respectively.}

\section*{Acknowledgments}
This work was supported by the Office of Naval Research (ONR) under award number UWIS-0000003591. The authors are grateful to Behzad Kamgar-Parsi for his support and encouragement. The authors thank Mohit Gupta and Jeremy Klotz for their technical feedback, and Aubrey Toland for discussions on engineered materials. The authors are also grateful to Hannah Fox, Sidharth Sharma, Joel Salzman, and Pranav Sukumar for help with collecting the Matador dataset, and Jessica Zhang and Lulu Wang for assistance refining the Matador website.

{
  \small
  \bibliographystyle{ieeenat_fullname}
  \bibliography{main}
}


\clearpage

\setcounter{page}{1}
\renewcommand{\thesection}{S\arabic{section}}
\renewcommand{\thefigure}{S\arabic{figure}}
\renewcommand{\thetable}{S\arabic{table}}
\renewcommand{\theequation}{S\arabic{equation}}
\setcounter{section}{0}
\setcounter{figure}{0}
\setcounter{table}{0}
\setcounter{equation}{0}
\setcounter{footnote}{0}

\maketitlesupplementary

\section{The Dearth of Material Image Data}
While material recognition as a line of research has existed for over 60 years, there exist only a few dozen widely available datasets. This is compared to thousands of datasets targeted at object recognition, segmentation, \etc, which can be more easily aggregated from online images and annotated by the layperson without any additional information. These two factors make the task of creating material image datasets very challenging. Nevertheless, there exists a need in the space of material recognition for not only more, but much larger datasets.

Beyond material recognition, there is rich literature on adjacent tasks that are highly correlated with recognizing materials. Several prior works introduce datasets for texture recognition~\citesupp{supp:cimpoiDescribingTexturesWild2014}, material segmentation~\citesupp{supp:bellOpenSurfacesRichlyAnnotated2013,supp:caiRGBRoadScene2022,supp:liangMultimodalMaterialSegmentation2022}, texture segmentation~\citesupp{supp:schwartzRecognizingMaterialProperties2020}, and estimation of the bidirectional texture function (BTF) or bidirectional reflectance function (BRDF)~\citesupp{supp:weinmannMaterialClassificationBased2014,supp:danaReflectanceTextureRealworld1999,supp:sattlerEfficientRealisticVisualization}. Even so, such datasets are often limited in their range of material categories. For example, while BRDF/BTF datasets may contain tens of thousands of images spread across a reasonable range of classes (usually between 10 to 40), typically, each class only has a few unique physical instances per class. Compare this to ImageNet \citesupp{supp:dengImageNetLargescaleHierarchical2009} for object recognition, which contains 1,000 classes with equally many instances per class. As a result, material recognition datasets have limited specificity with such few classes, and given the limited intraclass diversity, learned models may not generalize well to real-world tasks. We contribute \emph{Matador}, introduced in \cref{sec:dataset} and further detailed in \cref{sec:sup-datasets}, as a step towards overcoming this hurdle.

\section{Matador Dataset Details}\label{sec:sup-datasets}
\begin{figure}
  \centering
  \includegraphics[width=\linewidth]{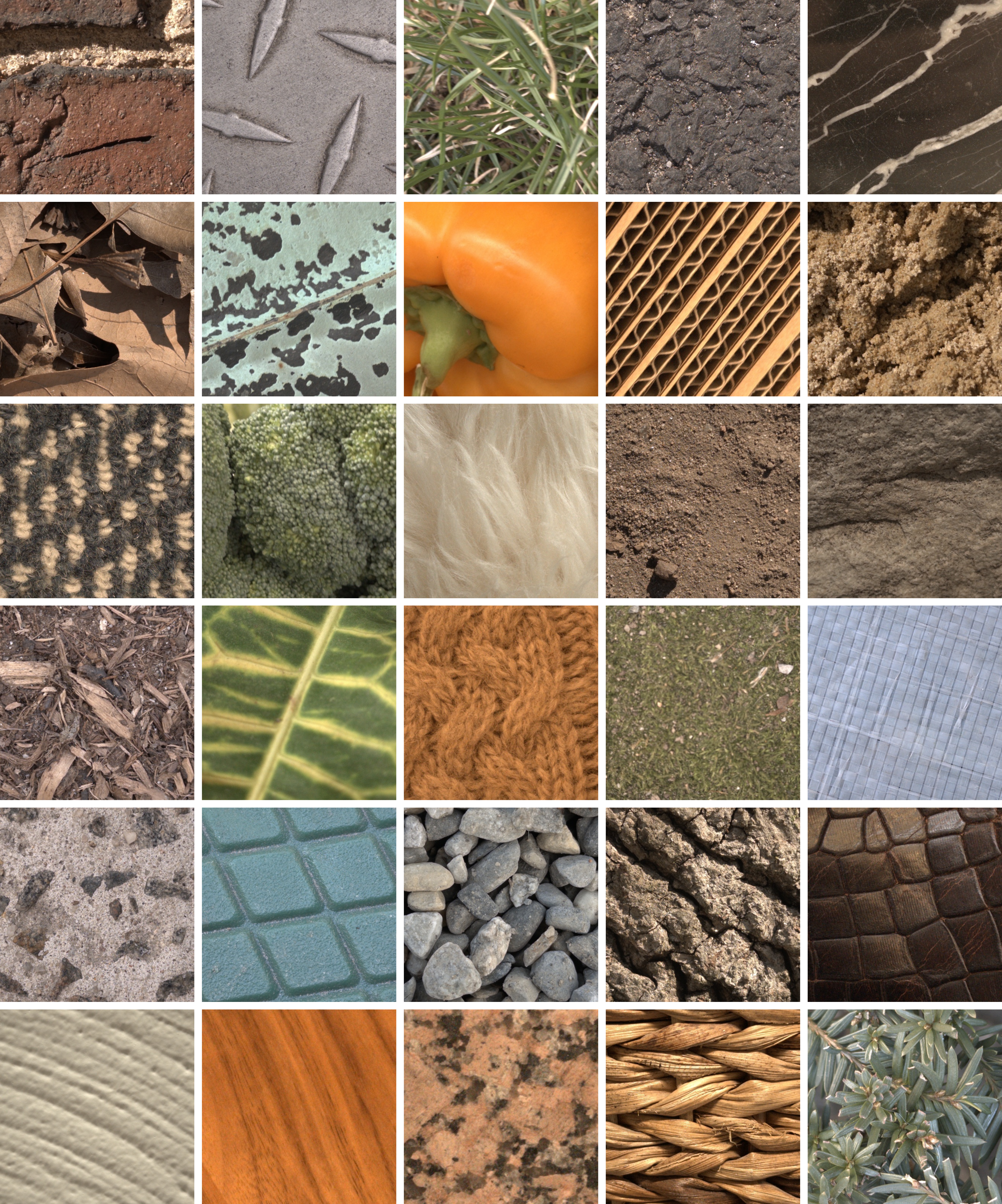}
  \vspace{-5.5mm}
  \caption{
    \textbf{\emph{Matador} Local Appearance Examples.} Several examples of the local appearance images in the \emph{Matador} dataset. These images are captured at a close distance (typically around 20-30 \unit{\cm} away), and ``local appearance'' denotes that the target material fills the majority of the frame. More examples can be examined using the interactive viewer on the \href{https://cave.cs.columbia.edu/repository/Matador}{\emph{Matador} webpage}.
  }\label{fig:matador-mosaic}
  \vspace{-0.2in}
\end{figure}

A subset of the most recent material recognition datasets used as benchmarks in this work is presented in \cref{tab:existing-datasets}. We compare our \emph{Matador} dataset and the subset \emph{Matador-C1} (see \cref{sec:sup-matadorc1}) to these datasets. We find \emph{Matador} contains higher interclass and intraclass diversity compared to existing datasets, and additionally incorporates new datatypes. The dataset is available on the \href{https://cave.cs.columbia.edu/repository/Matador}{\emph{Matador} webpage}, and representative samples are shown in \cref{fig:matador-mosaic}.

\begin{table*}
  \footnotesize
  \centering
  \caption{
    \textbf{Statistics on Datasets Used in This Work.} We compare the size and diversity of each benchmark dataset with the \emph{Matador} dataset, which contains a high number of material classes and instances per class (interclass and intraclass diversity). We refer the interested reader to \protect\citetsupp{supp:liuBoWCNNTwo2019} for metrics on older and adjacent datasets. Note that ``\emph{Matador} (raw)'' denotes the unprocessed version of the dataset, with images that are neither undistorted, demosaiced, nor cropped to a relevant region of interest.
  }\label{tab:existing-datasets}
  \vspace{-2.5mm}
  \begin{tabular}{llcccccc}
    \toprule
    Dataset                & Authors                                                                & Year & Classes & Images & Resolution~[px\textsuperscript{2}] & Avg.\ Instances per Class & Image Datatypes            \\
    \midrule
    KTH-TIPS-2b            & \protect\citetsupp{supp:caputoClassspecificMaterialCategorisation2005} & 2005 & 11      & 4,752  & 200\(\times\)200                   & 4                         & Appearance                 \\
    FMD                    & \protect\citetsupp{supp:sharanRecognizingMaterialsUsing2013}           & 2013 & 10      & 1,000  & 512\(\times\)384                   & 100                       & Appearance                 \\
    GTOS                   & \protect\citetsupp{supp:xueDifferentialAngularImaging2017}             & 2017 & 40      & 34,243 & 240\(\times\)240                   & \(\sim\)15                & Appearance, Angular        \\
    GTOS-Mobile            & \protect\citetsupp{supp:xueDeepTextureManifold2018}                    & 2018 & 31      & 6,066  & 455\(\times\)256                   & \(\sim\)5                 & Appearance                 \\
    \midrule
    \textit{Matador}       & Ours                                                                   & 2025 & 57      & 7,238  & 512\(\times\)512                   & \(\sim\)126               & Appearance, Depth, Context \\
    \textit{Matador-C1}    & Ours                                                                   & 2025 & 37      & 6,614  & 512\(\times\)512                   & \(\sim\)179               & Appearance, Depth, Context \\
    \textit{Matador} (raw) & Ours                                                                   & 2025 & 57      & 7,238  & 3024\(\times\)4032                 & \(\sim\)126               & Appearance, Depth, Context \\
    \bottomrule
  \end{tabular}
  \vspace{-0.1in}
\end{table*}

\subsection{Acquisition Details}
Here we provide further details about how the \emph{Matador} dataset was constructed. As discussed in \cref{sec:dataset}, we collected in-the-wild material images to populate the taxonomy with image data. We used an Apple iPhone 15 Pro Max to do this, of which we utilized three of its sensors: the wide-angle camera, lidar (which is registered with the wide-angle camera), and ultrawide-angle camera. We first take a close-up image of the material's local appearance using the wide-angle camera. Simultaneously, a lidar scan of the material's local surface structure is captured. We then take a step back to capture the material in the context of its
surroundings using the ultrawide-angle camera. Between these two captures, we record the motion of the
phone using the inertial measurement unit (IMU). The complete sample of a material thus contains the following raw data:
\begin{enumerate}
  \item \textbf{Local Appearance Image.} The image is 12~MP (3024\(\times\)4032~px\textsuperscript{2}) with a 74\textdegree~FOV from the wide-angle camera. Since the 48~MP wide-angle camera used to capture local appearance has a Quad-Bayer color filter array (for high dynamic range imaging in a single shot), the resolution after capture is reduced to 12~MP. The image is recorded as a 12-bit Bayer raw DNG.

  \item \textbf{Depth Map of the Surface Structure.} The image is a 100~\unit{points \per degree \squared} lidar depth map registered with the local appearance image. It has an equivalent FOV to the local appearance image, but when projected into two dimensions, the depth map is roughly a fifth of the resolution. Thus, it must be resampled to approximate per-pixel depth in the local appearance. The depth map is recorded as a 32-bit float binary file.

  \item \textbf{Global Context Image.} The image is 12~MP (3024\(\times\)4032~px\textsuperscript{2}) with a 104\textdegree~FOV from the ultrawide-angle camera. It is recorded as a 12-bit Bayer raw DNG.

  \item \textbf{Motion.} The IMU data contains accelerometer measurements in the time between the local appearance image capture and the global context image capture. Measurements are sampled at 100~\unit{\Hz} -- the maximum for the iPhone 15 Pro Max. The motion is recorded in a timestamped text file.

  \item \textbf{Metadata.} The metadata records camera intrinsics and extrinsics (from the wide-angle to ultrawide-angle camera) for each capture, as reported by the manufacturer. In addition, typical EXIF data is also recorded that includes the camera settings for each capture (\eg, ISO, exposure time, and suggested white balance gains).
\end{enumerate}
We record Bayer raw images to eliminate compression artifacts that could alter a material's appearance. The above datatypes represent the raw dataset. We developed an iOS application to complete this capture process; it can also be found on the \href{https://cave.cs.columbia.edu/repository/Matador}{\emph{Matador} webpage}. Note that in Swift, the wide-angle camera and lidar correspond to the \texttt{builtInLiDARDepthCamera}, and the ultrawide-angle camera corresponds to the \texttt{builtInUltraWideCamera}.

\subsection{Processing Raw Data into Matador}\label{sec:sup-raw2matador}
To process the raw data into the \emph{Matador} dataset, we first demosaic and undistort all images. Resampling is applied to the depth map to match the resolution of the local appearance image. We then define an approximately 5\(\times\)5~cm\textsuperscript{2} region of interest (using depth for scale) in the local appearance containing the material being captured and excluding extraneous information. The final appearance image is an sRGB 16-bit unsigned integer TIFF at 512\(\times\)512~px\textsuperscript{2}, and the corresponding depth map is a 512\(\times\)512~px\textsuperscript{2} 32-bit float TIFF. When resizing images, a Mitchell-Netravali filter is used for enlarging and a Lanczos filter is used for shrinking.\footnote{The merits of various resampling filters are discussed at length in the ImageMagick~\citesupp{supp:imagemagick} usage guide: \href{https://usage.imagemagick.org/filter/}{https://usage.imagemagick.org/filter/}.}
The context image remains 3024\(\times\)4032~px\textsuperscript{2} but is now an sRGB 16-bit unsigned integer TIFF. Motion data and metadata are YAMLs. We additionally correct for vignetting using manufacturer-provided gains (see the \href{https://helpx.adobe.com/content/dam/help/en/photoshop/pdf/DNG_Spec_1_7_1_0.pdf#page=110.42}{Adobe DNG specification}), which is only noticeable in the full-resolution context images.

\subsection{Formal Treatment of Novel View Rendering}
We now formalize the rendering process described in \cref{sec:dataset} used to generate novel views from \emph{Matador}, which are then utilized during model training. While this rendering process is standard, its application to material recognition is novel, and we detail the effects of each augmentation parameter below for completeness.

Novel view generation is motivated by the fact that the visual appearance of materials can vary wildly depending on the scale they take on in an image. By generating novel views, a model trained on both the real and rendered datasets may improve in generalization to real-world settings. This is supported by the out-of-distribution (OOD) results in \cref{fig:matador-accuracy:benchmarks}. A qualitative look at this can be seen in \cref{fig:probing}, where materials are recognized at distances well beyond those captured in the local appearance images of \emph{Matador} that are used for training.

Since the 3D structure and position of the material are known, we can alter camera pose and imaging characteristics to simulate images produced by different cameras and viewpoints. We assert the true image irradiance $E$ is well-sampled by the local appearance image. With the local appearance and structure, we create a mesh of the material surface where each face represents the scene radiance $L$ -- assuming Lambertian reflectance. Spatial transformations $H$ (magnification and orientation) can then be applied to the mesh to change the pose of the imaged material relative to the observing camera. The novel view irradiance $E'$ is proportional to the scene radiance $L$, and we retrieve $E'$ through raytracing~\citesupp{supp:jakobDRJITJustintimeCompiler2022} accounting for occlusions, perspective effects, lens aberrations (\eg, defocus), and attenuation (from the aperture, visibility, \etc). The irradiance of a novel view is then:
\begin{align}
  E'_b(\mathbf{x}) & = \frac{1}{\text{Area}(P)}\int_{\mathbf{p} \in P}L\left(\mathbf{p} + t (F(\mathbf{x}) - \mathbf{p})\right)\,d\mathbf{p}, \label{eq:novel-irradiance-continuous}
\end{align}
where $t$ is the length of the ray, and,
\begin{align}
  F(\mathbf{x}) & = \mathbf{c} + \frac{d_f}{||\mathbf{x}-\mathbf{c}||}\,(\mathbf{x} - \mathbf{c}),
\end{align}
maps a sensor coordinate $\mathbf{x}$ to a point on the focal plane relative to the piercing point $\mathbf{c}$. The integral is over each point $\mathbf{p}$ on the effective pupil $P$ with full transmission, equivalent to the aperture under a thin-lens model. Spatially-varying blur is then introduced to the novel view by varying the focus distance $d_f$ and size of the pupil $P$.

The irradiance in \cref{eq:novel-irradiance-continuous} is continuous and must be sampled to create an image. We first convolve with a box filter $p$ to average over the active area of each pixel:
\begin{align}
  E'_a(\mathbf{x}) & = E'_b(\mathbf{x}) * p(\mathbf{x}).
\end{align}
The density of rays being traced is then equivalent to sampling with a pulse train with period $\Delta$:
\begin{align}
  E'_s(\mathbf{x}) & = E'_a(\mathbf{x}) \; \boldsymbol{\cdot} \sum_{\mathbf{k} \in \mathbb{Z}^2} \delta (\mathbf{x}-\Delta \mathbf{k}).
\end{align}
We additionally model photon and sensor noise with a zero-mean random variable $n$~\citesupp{supp:hasinoffNoiseoptimalCaptureHigh2010}:
\begin{align}
  E'_n(\mathbf{x}) & = E'_s(\mathbf{x}) + n(\mathbf{x}).
\end{align}
Finally, the result is discretized by the pixel pitch $T$ to produce an image:
\begin{align}
  E'[\mathbf{u}] & = E'_n(\mathbf{u}T).
\end{align}
The resulting views are cropped to fill the entire frame and resized to a standard 512\(\times\)512 px\textsuperscript{2} size. As in \cref{sec:sup-raw2matador}, when resizing images, a Mitchell-Netravali filter is used for enlarging and a Lanczos filter is used for shrinking.

By altering the pose of the material (with $H$), the defocus of the lens (with $d_f$ and $P$), the pixel size $p$, the sensor noise $n$, and the pitch $T$, we can simulate how an arbitrary camera would render the material from any distance and orientation. This process is applied to all raw samples in the \emph{Matador} dataset to obtain a larger and more diverse set of training images. Numerous novel views are rendered per raw sample, faithfully depicting how a given material will look under a wide range of imaging conditions.

\subsection{Generalization Test Set Details}\label{sec:supp-generalization-test-set}
Novel views supplement the captured local appearance images used for model training. In addition to local appearance images used for testing, we additionally create a second evaluation set, which we refer to as the out-of-distribution (OOD) test set. As discussed in \cref{sec:dataset}, the OOD test set is intended to evaluate the effect of generated novel views on model generalization to real-world imaging conditions (such as further viewpoints, harsh blur, and viewing directions that are not frontoparallel). Here we further detail how this test set was created.

In each raw local appearance image, a square region of interest (ROI) is used to define a relevant patch for model training and testing. This is done to remove any context that might be present in the full-resolution raw local appearance image, and it is typically 5\(\times\)5~cm\textsuperscript{2}. Using this same ROI and the registered depth map, we map the four corners of the ROI into three dimensions through unprojection. We then apply a Kalman filter to the accelerometer measurements between the appearance and context captures, and use the resulting spatial transformation to transform the 3D ROI corners into the wide-angle camera's frame of reference at the time of the context image capture. Using manufacturer-provided camera extrinsics, we then transform the 3D ROI corners into the coordinate frame of the ultrawide-angle camera and project them (incorporating differences in camera intrinsics) to get an ROI of similar scale in the context image. This ROI is extracted and resized to 512\(\times\)512~px\textsuperscript{2}.

\begin{figure}
  \centering
  \footnotesize
  \setlength{\tabcolsep}{1pt}
  \begin{tabular}{m{2.5mm}p{\dimexpr0.98\columnwidth-2.5mm\relax}}
                                                     &
    \begin{minipage}{\linewidth}
      \begin{subfigure}[b]{0.24\linewidth}
        \caption*{\centering Shrub}
      \end{subfigure}
      \hfill
      \begin{subfigure}[b]{0.24\linewidth}
        \caption*{\centering Silk}
      \end{subfigure}
      \hfill
      \begin{subfigure}[b]{0.24\linewidth}
        \caption*{\centering Fruit}
      \end{subfigure}
      \hfill
      \begin{subfigure}[b]{0.24\linewidth}
        \caption*{\centering Dirt}
      \end{subfigure}
    \end{minipage}                      \\

    \centering \rotatebox{90}{\textbf{Original ROI}} &
    \begin{minipage}{\linewidth}
      \centering
      \begin{subfigure}[b]{0.24\linewidth}
        \includegraphics[width=\linewidth, height=\linewidth]{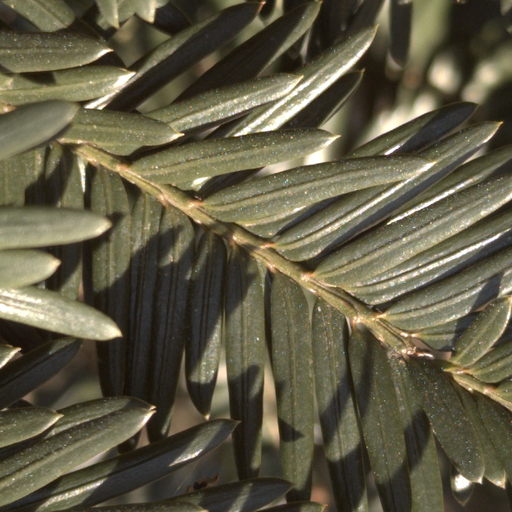}
      \end{subfigure}
      \hfill
      \begin{subfigure}[b]{0.24\linewidth}
        \includegraphics[width=\linewidth, height=\linewidth]{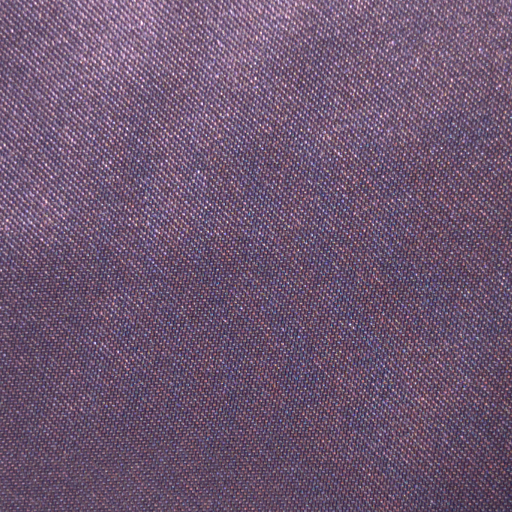}
      \end{subfigure}
      \hfill
      \begin{subfigure}[b]{0.24\linewidth}
        \includegraphics[width=\linewidth, height=\linewidth]{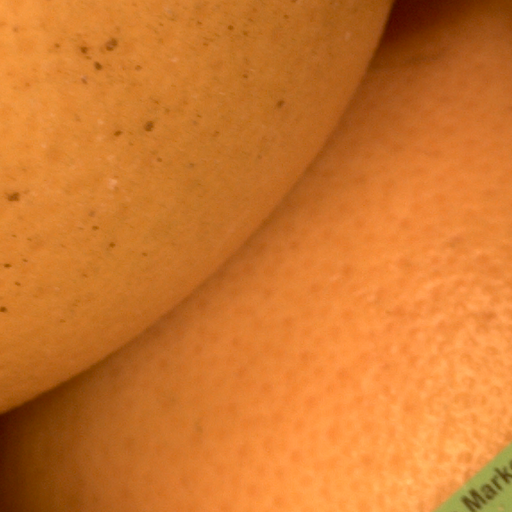}
      \end{subfigure}
      \hfill
      \begin{subfigure}[b]{0.24\linewidth}
        \includegraphics[width=\linewidth, height=\linewidth]{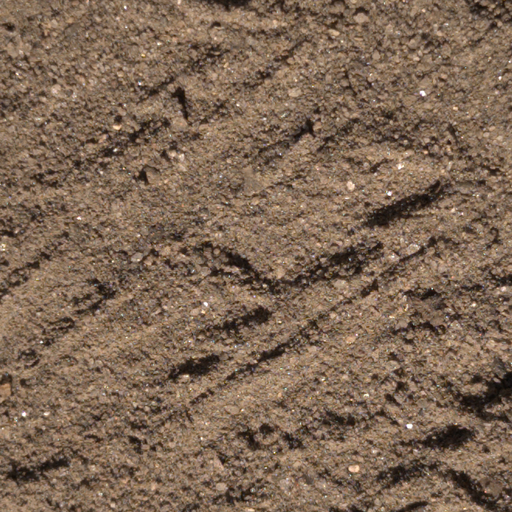}
      \end{subfigure}
      \vspace{2pt}
    \end{minipage}                      \\

    \centering \rotatebox{90}{\textbf{Context ROI}}  &
    \begin{minipage}{\linewidth}
      \centering
      \begin{subfigure}[b]{0.24\linewidth}
        \includegraphics[width=\linewidth, height=\linewidth]{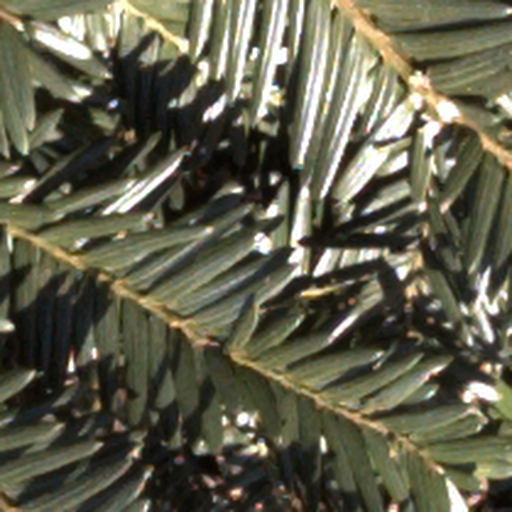}
      \end{subfigure}
      \hfill
      \begin{subfigure}[b]{0.24\linewidth}
        \includegraphics[width=\linewidth, height=\linewidth]{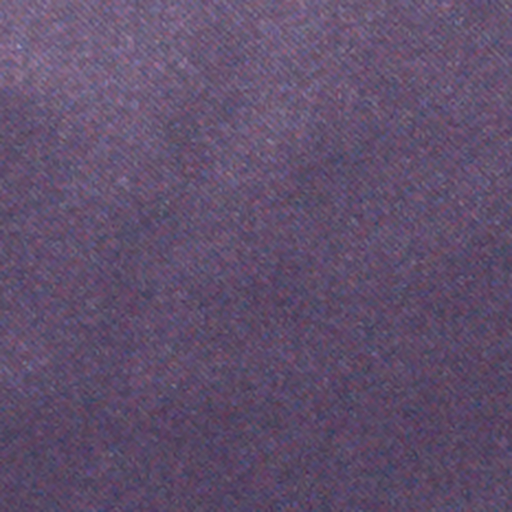}
      \end{subfigure}
      \hfill
      \begin{subfigure}[b]{0.24\linewidth}
        \includegraphics[width=\linewidth, height=\linewidth]{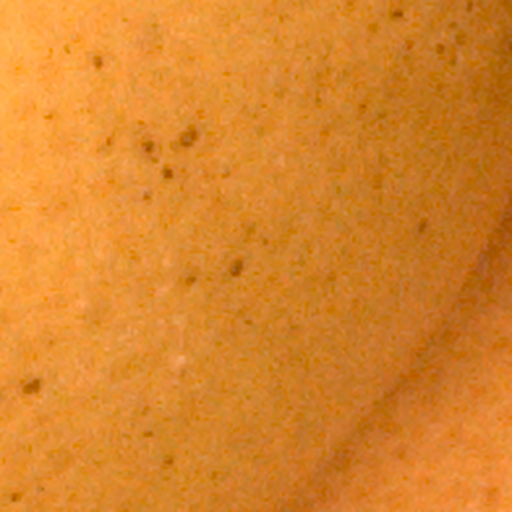}
      \end{subfigure}
      \hfill
      \begin{subfigure}[b]{0.24\linewidth}
        \includegraphics[width=\linewidth, height=\linewidth]{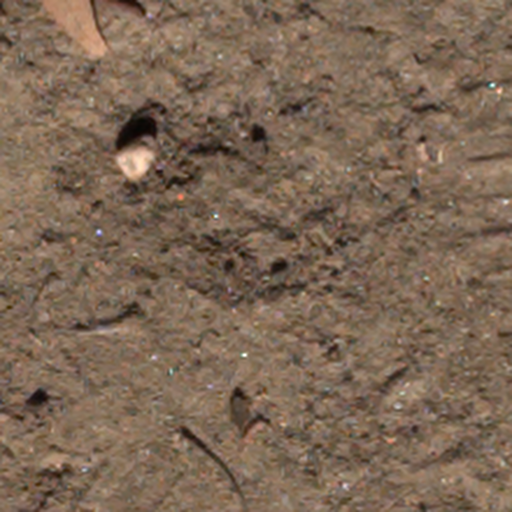}
      \end{subfigure}
      \vspace{2pt}
    \end{minipage}                      \\
  \end{tabular}

  \vspace{-2mm}
  \caption{
    \textbf{Evaluating Generalization.} Examples of the appearance image (top) and patch cropped from the context image (bottom) for \emph{Matador} samples. The latter is used to create an out-of-distribution test set to evaluate real-world model performance (see \cref{sec:supp-generalization-test-set}). Compared to the original ROIs used for model training, the context ROIs come from a different camera, a more distant viewpoint, and capture a different area of the material sample. As a result, they generally exhibit reduced detail, have a slight blur, and appear flatter. Zoom in to best see the fine texture differences between the images.
  }\label{fig:generalization-test-set}
  \vspace{-0.2in}
\end{figure}

Due to noise in the accelerometer measurements, this is not an exact mapping between the appearance and context ROIs. Instead, the context ROI samples a different location on the target material instance. We manually verify this and that the resulting patches capture the correct material. Since they capture a different location on the material using a more distant viewpoint, different lens, and different image sensor, these patches comprise the OOD test set that simulates realistic real-world capture conditions (see \cref{fig:generalization-test-set}).

\begin{figure*}[h]
  \centering
  \begin{minipage}[b]{0.47\linewidth}
    \centering
    \includegraphics[width=\linewidth]{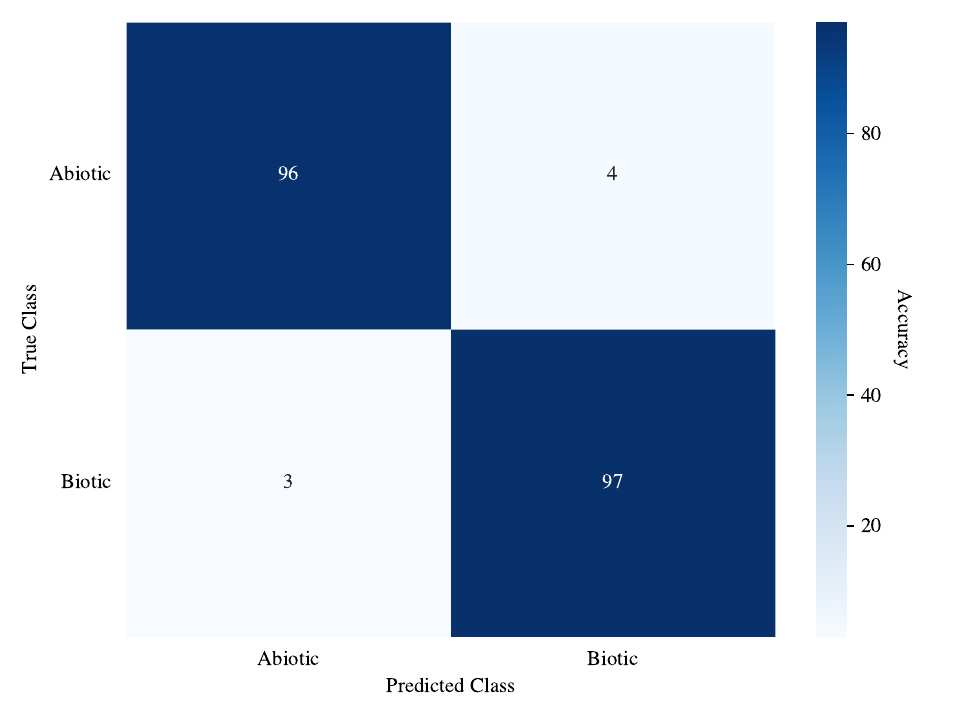}
    \subcaption{State}
  \end{minipage}
  \hspace{0.25cm}
  \begin{minipage}[b]{0.47\linewidth}
    \centering
    \includegraphics[width=\linewidth]{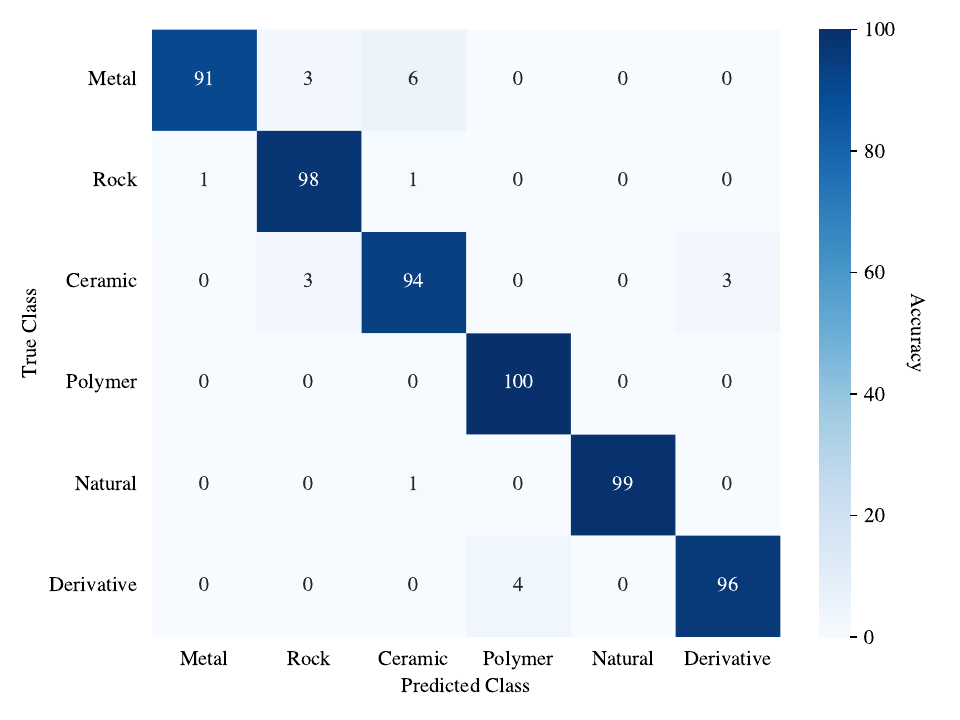}
    \subcaption{Composition}
  \end{minipage}

  \vspace{0.25cm} 

  \begin{minipage}[b]{0.47\linewidth}
    \centering
    \includegraphics[width=\linewidth]{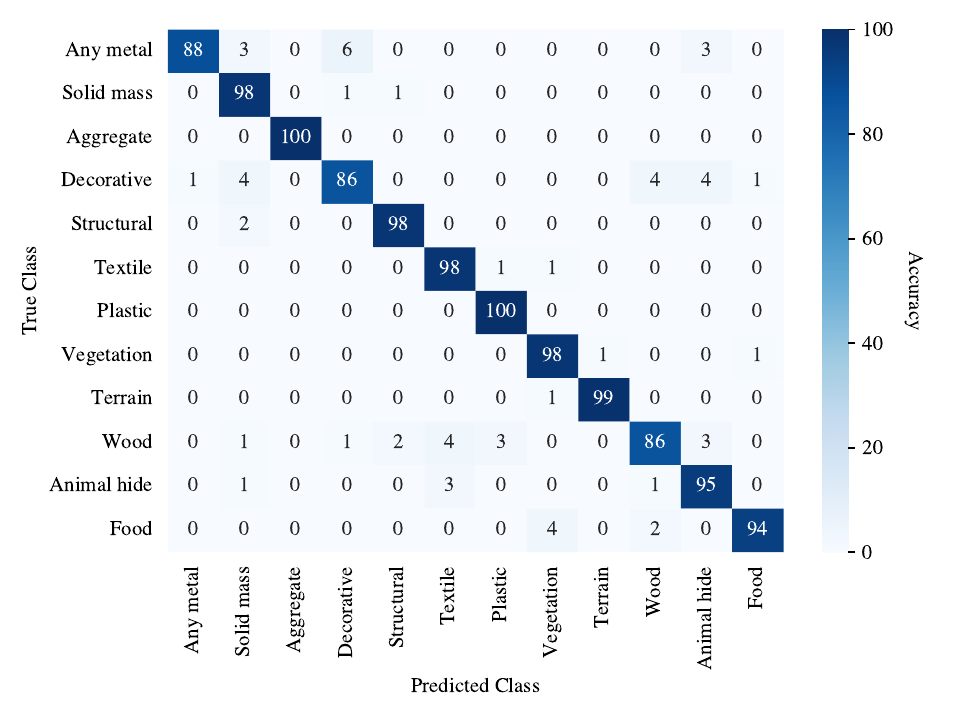}
    \subcaption{Form}
  \end{minipage}
  \hspace{0.25cm}
  \begin{minipage}[b]{0.47\linewidth}
    \centering
    \includegraphics[width=\linewidth]{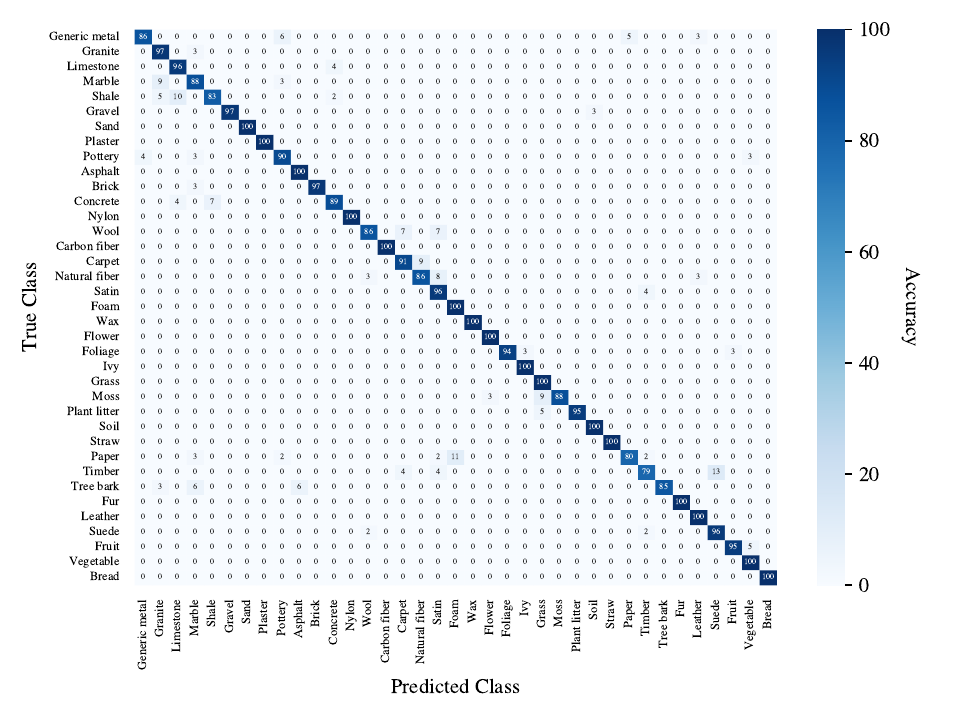}
    \subcaption{Material}
  \end{minipage}
  \vspace{-0.025in}
  \caption{
    \textbf{Class Accuracy by Taxonomy Level.} Top-1 accuracy of our model on \emph{Matador-C1} (Color), per class for each level in our taxonomy. Plots (a-d) are arranged in order of increasing specificity, referring to the levels shown in \cref{fig:taxonomy-tree}. Values are rounded to the nearest integer percentage using the largest remainder method. By framing material recognition as a hierarchical learning problem, we are able to accurately recognize materials at multiple levels of granularity. Note that since all metals are consolidated into a single class in \emph{Matador-C1}, the classes ``Any metal'' in~(c) and ``Generic metal'' in~(d) are equivalent to ``Metal'' in~(b). Nevertheless, we provide additional evaluations at the ``Form'' and ``Material'' levels for completeness.
  }\label{fig:eval-class-confusion-matrices}
  \vspace{-0.15in}
\end{figure*}

\subsection{Consolidating Matador into Matador-C1}\label{sec:sup-matadorc1}
For the performance metrics in \cref{fig:matador-accuracy:benchmarks}, we consolidated the \emph{Matador} dataset from all 57 categories present in the taxonomy into a smaller set with 37 categories -- we refer to the consolidated dataset as \emph{Matador-C1}. In the process, several categories were omitted (thermoplastic, thermoset, elastomer, paint, and glass) that we deemed to have insufficient texture to recognize solely from their local appearance. For instance, glass is featureless and its appearance is dominated by reflections, and paint, being thin, essentially takes on the surface structure of the material it sits on. For similar reasons, we combined other categories into a single one: \{aluminum, steel, brass, iron, bronze, copper\} are referred to as ``generic metal'', \{stoneware, terracotta, porcelain\} as ``pottery'', \{dirt, soil\} as ``soil'', \{shrub, foliage\} as ``foliage'', \{sandstone, shale\} as ``shale'', \{marble, quartz\} as ``marble'', \{polyester, silk\} as ``satin'', \{cotton, linen\} as ``natural fiber'', \{cardboard, paper\} as ``paper'', and \{cement, concrete\} as ``concrete''. The visual features of such materials are highly dependent on surface finish and patina, and thus recognition requires knowing context.

For many of the materials omitted or combined in \emph{Matador-C1}, it would be more reasonable to correctly recognize them with the inclusion of global context and even depth images. As the focus of this paper is recognition from local appearance, we did not explore this avenue. However, these data types remain available in the complete \emph{Matador} dataset for future work to leverage.

\section{Model Training and Evaluation Details}\label{sec:sup-evaluation}
We now discuss the implementation details of model construction, training, and evaluation. We use a ResNet50~\citesupp{supp:heDeepResidualLearning2016} as the image encoder and configure the hierarchical graph attention network defined in \cref{sec:method} with the following setup:
{
\begin{center}
  \begin{tabular*}{\linewidth}{p{0.42\linewidth} p{0.5\linewidth}@{}}
    \textbf{Input dim:} 1024 & \textbf{Layers:} 2            \\
    \textbf{Hidden dim:} 512 & \textbf{Attention heads:} 1   \\
    \textbf{Output dim:} 256 & \textbf{Pool:} Global Average \\
  \end{tabular*}
\end{center}
}
\noindent Prototype embeddings are then created for each representative class in the dataset with dimensionality equal to the GNN input dimension. The entire model is then trained end-to-end with the following hyperparameters:
{
\begin{center}
  \begin{tabular*}{\linewidth}{p{0.42\linewidth} p{0.5\linewidth}@{}}
    \textbf{Batch size:} 400           & \textbf{LR Schedule:} Cos.\ anneal. \\
    \textbf{Epochs:} 100               & \textbf{Weight decay:} \num{5e-4}   \\
    \textbf{Learning rate:} \num{1e-4} & \textbf{Sampling:} Stratified       \\
  \end{tabular*}
\end{center}
}
\noindent With a ResNet50 image encoder and the \emph{Matador-C1} classes, the model contains 28.0~M parameters. Training is performed on NVIDIA A6000 Ada GPUs, and if training does not use rendered novel views, it can complete training in under 30 minutes on a single GPU.

\begin{table}
  \centering
  \footnotesize
  \begin{threeparttable}
    \caption{
      \textbf{Additional Performance Metrics on \emph{Matador-C1}.} Comparison with existing methods. The best overall results are highlighted in bold, and the second best are underlined. Extending \cref{fig:matador-accuracy:benchmarks}, we add evaluations on grayscale images as well as when only finetuning the classifier head of competing material recognition models. All models use a ResNet50 backbone.
      \vspace{-0.1in}
    }\label{tab:additional-matador-accuracy}
    \setlength{\tabcolsep}{9pt}
    \begin{tabular}{lccc}
      \toprule
                                                                        & \multicolumn{3}{c}{Top-1 Accuracy \(\uparrow\)}                                               \\
      \cmidrule(l{\tabcolsep}r{\tabcolsep}){2-4}
      Method                                                            & Grayscale                                       & Color                & Out-of-Distribution  \\
      \midrule
      \multicolumn{4}{l}{\emph{Material Recognition Models (Finetuned Classifier Head)}} \vspace{1pt}                                                                   \\
      \, DeepTEN~\protect\citesupp{supp:zhangDeepTENTexture2017}        & \(67.1\)                                        & \(54.0\)             & \(52.5\)             \\
      \, DEPNet~\protect\citesupp{supp:xueDeepTextureManifold2018}      & \(69.6\)                                        & \(76.8\)             & \(64.3\)             \\
      \, FRP~\protect\citesupp{supp:florindoFractalPoolingNew2024}      & \(74.2\)                                        & \(68.9\)             & \(58.4\)             \\
      \, MSLac~\protect\citesupp{supp:mohanLacunarityPoolingLayers2024} & \(74.0\)                                        & \(79.1\)             & \(65.7\)             \\
      \midrule
      \multicolumn{4}{l}{\emph{Material Recognition Models (Finetuned End-to-End)}} \vspace{1pt}                                                                        \\
      \, DeepTEN~\protect\citesupp{supp:zhangDeepTENTexture2017}        & \(80.3\)                                        & \(88.8\)             & \(61.5\)             \\
      \, DEPNet~\protect\citesupp{supp:xueDeepTextureManifold2018}      & \(84.2\)                                        & \(87.6\)             & \underline{\(76.1\)} \\
      \, FRP~\protect\citesupp{supp:florindoFractalPoolingNew2024}      & \(84.6\)                                        & \underline{\(89.4\)} & \(71.0\)             \\
      \, MSLac~\protect\citesupp{supp:mohanLacunarityPoolingLayers2024} & \underline{\(84.7\)}                            & \(88.5\)             & \(75.4\)             \\
      \midrule
      \multicolumn{4}{l}{\emph{\textbf{Hierarchical Material Recognition Models}}} \vspace{1pt}                                                                         \\
      \, Ours                                                           & \(\mathbf{87.5}\)                               & \(\mathbf{94.1}\)    & \(\mathbf{82.9}\)    \\
      \bottomrule
    \end{tabular}
  \end{threeparttable}
  \vspace{-0.1in}
\end{table}

We present the performance of our hierarchical graph attention network on \emph{Matador-C1} (Color) for each class in each level of the taxonomy in \cref{fig:eval-class-confusion-matrices}. Through hierarchical learning, our model exhibits high accuracy at each level. A hierarchical model allows recognition to appropriate levels of specificity depending on the application task, with increasing recognition performance at coarser levels.

In \cref{tab:additional-matador-accuracy}, we present additional evaluations on \emph{Matador-C1}. We evaluate our model and competing methods on color and grayscale images, finetuning competing methods in two ways. In each case, our hierarchical model achieves state-of-the-art classification accuracy.

\section{Material Properties}\label{sec:sup-material-properties}
To enable intelligent systems operating in the physical world, we aggregated a table of mechanical properties (\cref{tab:mechanical-properties}) containing the materials of the taxonomy described in \cref{sec:taxonomy}. After visual recognition of a material, the properties present in the table can be used to plan interactions within the environment. For example, if a material is deformable, it will require different handling than if it were rigid. To this end, \cref{tab:mechanical-properties} is composed of approximate ranges of material density, surface roughness, elasticity, and strength. Its content is retrieved from common textbooks, engineering specification sheets, material databases, and academic publications. These material properties vary in precision -- engineered materials (\eg, metals, ceramics, and plastics) are studied extensively and thus their property ranges are well known. For natural materials (\eg, foliage, moss, and grass), rough estimates are provided.

\begin{table*}
  \footnotesize
  \centering
  \setlength{\tabcolsep}{3.75pt}
  \caption{
    \textbf{Approximate Mechanical Properties of Common Materials.} Sourced from various textbooks, handbooks, and articles on material properties, the present table contains the mechanical properties for each material in our taxonomy. While rough estimates for many categories, such information could be useful to make judgments about whether a recognized material is, \eg, light or heavy, smooth or rough, deformable or rigid, and fragile or strong. Note: a loose approximation of shear strength can also be obtained by multiplying the tensile strength by \(1/2\) for brittle materials and \(3/5\) for ductile materials.
    \vspace{-0.1in}
  }\label{tab:mechanical-properties}
  \rowcolors{2}{white}{gray!15}
  \begin{tabular}{m{35mm}
    >{\centering}m{19mm}
    >{\centering}m{22mm}
    >{\centering}m{20mm}
    >{\centering}m{20mm}
    >{\centering}m{20mm}
    >{\centering\arraybackslash}m{20mm}}
    \toprule
    \textbf{Material}               & Density [\unit{\kg\per\meter\cubed}] & Surface Roughness [\unit{\um}] & Young's Modulus [\unit{\GPa}] & Yield Strength [\unit{\MPa}] & Tensile Strength [\unit{\MPa}] & Poisson's Ratio \\
    \midrule
    Iron (pure/cast forms)          & 7150--7870                           & 0.1--50                        & 100--210                      & 120--200                     & 130--210                       & 0.26--0.30      \\
    Steel (alloyed iron)            & 7850--8000                           & 0.1--10                        & 190--210                      & 250--1000                    & 400--2000                      & 0.27--0.30      \\
    Aluminum (alloys)               & 2700--2830                           & 0.2--10                        & 69--72                        & 30--400                      & 50--550                        & 0.32--0.35      \\
    Brass                           & 8400--8730                           & 0.2--5                         & 100--125                      & 180--250                     & 350--600                       & 0.33--0.36      \\
    Bronze                          & 7400--8920                           & 0.2--5                         & 96--120                       & 140--380                     & 240--586                       & 0.34--0.36      \\
    Copper                          & 8930--8960                           & 0.1--5                         & 110--130                      & 60--70                       & 200--220                       & 0.34--0.36      \\
    Granite                         & 2600--2700                           & 1--50                          & 50--70                        & --                           & 7--25                          & 0.20--0.30      \\
    Limestone                       & 2300--2700                           & 1--50                          & 15--55                        & --                           & 5--25                          & 0.20--0.30      \\
    Marble                          & 2400--2700                           & 0.5--5                         & 50--70                        & --                           & 7--20                          & 0.20--0.30      \\
    Sandstone                       & 2200--2800                           & 5--50                          & 1--20                         & --                           & 4--25                          & 0.10--0.25      \\
    Shale                           & 1770--2670                           & 5--50                          & 1--70                         & --                           & 2--10                          & 0.10--0.30      \\
    Quartz                          & 2600--2700                           & 0.05--5                        & 70--100                       & --                           & 10--30                         & 0.16--0.18      \\
    Dirt (dry, loose)               & 1000--1600                           & 50--2000                       & 0.01--0.2                     & --                           & 0--0.02                        & 0.20--0.40      \\
    Gravel (loose aggregate)        & 1500--1800                           & 2000--50000                    & 0.05--0.2                     & --                           & 0.2--0.6                       & 0.20--0.30      \\
    Sand (dry, loose)               & 1400--1700                           & 50--500                        & 0.01--0.07                    & --                           & 0.1--0.3                       & 0.20--0.46      \\
    Glass (soda-lime)               & 2500--2600                           & 0.05--1                        & 70--90                        & --                           & 30--90                         & 0.20--0.25      \\
    Plaster (gypsum)                & 600--1900                            & 1--10                          & 1--10                         & --                           & 2--6                           & 0.18--0.22      \\
    Porcelain                       & 2380--2450                           & 0.1--1                         & 50--74                        & --                           & 30--50                         & 0.17--0.25      \\
    Stoneware                       & 2000--2400                           & 0.5--5                         & 30--70                        & --                           & 20--40                         & 0.17--0.25      \\
    Terracotta (earthenware)        & 1600--1900                           & 5--50                          & 10--30                        & --                           & 2--5                           & 0.17--0.25      \\
    Asphalt (pavement)              & 2200--2400                           & 50--500                        & 3--11                         & --                           & 0.5--3                         & 0.20--0.35      \\
    Brick (clay)                    & 1600--2000                           & 5--50                          & 5--20                         & --                           & 2--5                           & 0.15--0.20      \\
    Cement (hardened paste)         & 1900--2200                           & 5--50                          & 10--30                        & --                           & 2--5                           & 0.20--0.30      \\
    Concrete (aggregate mix)        & 2200--2500                           & 10--200                        & 25--40                        & --                           & 2--5                           & 0.10--0.20      \\
    Cotton (textile)                & 1500--1600                           & 0.05--0.5                      & 2--8                          & --                           & 5--15                          & 0.30--0.40      \\
    Linen (textile)                 & 1400--1600                           & 0.1--1                         & 2--8                          & --                           & 20--60                         & 0.30--0.40      \\
    Nylon (textile)                 & 1110--1150                           & 0.002--0.006                   & 2--6                          & --                           & 200--500                       & 0.35--0.40      \\
    Polyester (PET, textile)        & 1330--1380                           & 0.004--0.006                   & 3--7                          & --                           & 200--500                       & 0.35--0.40      \\
    Silk (textile)                  & 1300--1360                           & 0.005--0.05                    & 5--10                         & --                           & 100--400                       & 0.30--0.40      \\
    Wool (textile)                  & 1200--1400                           & 0.8--3                         & 1--3                          & --                           & 30--150                        & 0.35--0.45      \\
    Carbon Fiber (textile/laminate) & 1550--1800                           & 0.007--10                      & 70--500                       & --                           & 1100--5000                     & 0.20--0.35      \\
    Carpet (fiber pile)             & 100--600                             & 500--5000                      & 0.001--0.01                   & --                           & 1--3                           & 0.30--0.40      \\
    Elastomer (rubber)              & 900--1200                            & 0.5--50                        & 0.003--0.01                   & --                           & 5--25                          & 0.48--0.50      \\
    Foam (polymer foam)             & 20--300                              & 0.05--26                       & 0.001--0.1                    & --                           & 0.05--5                        & 0.10--0.50      \\
    Paint (dry film)                & 1200--1500                           & 0.5--5                         & 0.5--3                        & --                           & 5--50                          & 0.30--0.40      \\
    Thermoplastic (bulk plastic)    & 900--1400                            & 0.2--5                         & 0.5--3                        & 20--70                       & 30--100                        & 0.32--0.35      \\
    Thermoset (rigid polymer)       & 1100--1500                           & 0.5--5                         & 2--5                          & 30--80                       & 40--80                         & 0.35--0.40      \\
    Wax                             & 900--970                             & 1--10                          & 0.03--0.08                    & --                           & 0.8--1.2                       & 0.30--0.35      \\
    Flower                          & 800--1000                            & 0.005--10                      & 0.005--0.02                   & --                           & 0.2--1                         & 0.30--0.40      \\
    Foliage                         & 600--1000                            & 5--25                          & 0.005--0.02                   & --                           & 0.1--0.3                       & 0.30--0.40      \\
    Ivy (vine)                      & 600--1000                            & 5--25                          & 0.1--5                        & --                           & 5--50                          & 0.30--0.40      \\
    Shrub (woody plant)             & 500--900                             & 5--25                          & 1--10                         & --                           & 40--100                        & 0.20--0.30      \\
    Grass (blade)                   & 800--1000                            & 5--25                          & 0.005--0.02                   & --                           & 0.1--0.5                       & 0.30--0.40      \\
    Moss                            & 50--300                              & 3--10                          & 0.0005--0.005                 & --                           & 0.1--0.7                       & 0.30--0.40      \\
    Plant Litter (dry leaves)       & 50--100                              & 5--25                          & 0.001--0.005                  & --                           & 0.01--0.1                      & 0.30--0.40      \\
    Soil (moist, packed)            & 1400--2000                           & 50--2000                       & 0.05--0.2                     & --                           & 0.02--0.4                      & 0.20--0.40      \\
    Straw (aggregate)               & 70--150                              & 100--1000                      & 1--5                          & --                           & 20--50                         & 0.30--0.40      \\
    Cardboard                       & 600--800                             & 10--100                        & 1--5                          & --                           & 20--80                         & 0.35--0.45      \\
    Paper (sheet)                   & 700--1200                            & 1--20                          & 2--5                          & --                           & 30--80                         & 0.30--0.40      \\
    Timber (wood)                   & 400--900                             & 5--50                          & 6--18                         & 25--140                      & 40--150                        & 0.25--0.40      \\
    Tree Bark (outer)               & 200--600                             & 200--2000                      & 0.01--0.1                     & --                           & 1--5                           & 0.30--0.40      \\
    Fur (pelt)                      & 1200--1400                           & 10--200                        & 1--3                          & --                           & 30--100                        & 0.35--0.45      \\
    Leather                         & 900--1000                            & 5--30                          & 0.2--0.5                      & --                           & 20--30                         & 0.40--0.45      \\
    Suede (split leather)           & 850--950                             & 10--50                         & 0.1--0.3                      & --                           & 15--30                         & 0.35--0.45      \\
    Fruit (fresh)                   & 900--1100                            & 0.3--15                        & 0.005--0.01                   & --                           & 0.1--0.3                       & 0.40--0.50      \\
    Vegetable (fresh)               & 800--1100                            & 0.4--14                        & 0.001--0.01                   & --                           & 0.05--0.5                      & 0.40--0.50      \\
    Bread (crumb)                   & 200--400                             & 500--2000                      & 0.00001--0.0002               & --                           & 0.001--0.01                    & 0.30--0.40      \\
    \bottomrule
  \end{tabular}
\end{table*}

{
  \small
  \bibliographystylesupp{ieeenat_fullname}
  \bibliographysupp{supp}
}

\end{document}